\documentclass[preprint,12pt,authoryear]{elsarticle}
\usepackage[a4paper,top=2cm,bottom=2cm,left=2.5cm,right=2.5cm,marginparwidth=1.75cm]{geometry}
\usepackage{amsmath}
\usepackage{graphicx} % Required for inserting images
\usepackage{makecell}
\usepackage{multirow}
\usepackage[table]{xcolor}
\usepackage[title]{appendix}
\usepackage{float}
\usepackage[section]{placeins}
\usepackage{amsmath}
\usepackage{bm}
\usepackage{paralist}
\usepackage{microtype}
\usepackage{algorithm}
\usepackage{algpseudocode}
\usepackage{url}
\usepackage{natbib}

\journal{a journal}

\begin{document}

\begin{frontmatter}

\title{Engineering an Efficient Object Tracker for Non-Linear Motion}

\author[1]{Momir Adžemović}\ead{pd222011@alas.matf.bg.ac.rs}\cortext[cor1]{Corresponding author}
\author[2]{Predrag Tadić}
\author[3]{Andrija Petrović}
\author[1]{Mladen Nikolić}

\affiliation[1]{organization={Department of Computer Science, Faculty of Mathematics, University of Belgrade},
addressline={Studentski trg 16}, 
city={Belgrade},
postcode={105104}, 
country={Serbia}}

\affiliation[2]{organization={Department of Signals and Systems, School of Electrical Engineering, University of Belgrade},
addressline={Bulevar kralja Aleksandra 73}, 
city={Belgrade},
postcode={11120}, 
country={Serbia}}

\affiliation[3]{organization={Center for Bussiness Decision Making, Faculty of Organizational sciences, University of Belgrade},
addressline={Jove Ilića 154}, 
city={Belgrade},
postcode={11010}, 
country={Serbia}}

\begin{abstract}

The goal of multi-object tracking is to detect and track all objects in a scene while maintaining unique identifiers for each, by associating their bounding boxes across video frames. This association relies on matching motion and appearance patterns of detected objects. This task is especially hard in case of scenarios involving dynamic and non-linear motion patterns. 
In this paper, we introduce DeepMoveSORT, a novel, carefully engineered multi-object tracker designed specifically for such scenarios. In addition to standard methods of appearance-based association, we improve motion-based association by employing deep learnable filters (instead of the most commonly used Kalman filter) and a rich set of newly proposed heuristics. 
Our improvements to motion-based association methods are severalfold. First, we propose a new transformer-based filter architecture, TransFilter, which uses an object's motion history for both motion prediction and noise filtering. We further enhance the filter's performance by careful handling of its motion history and accounting for camera motion. Second, we propose a set of heuristics that exploit cues from the position, shape, and confidence of detected bounding boxes to improve association performance. 
Our experimental evaluation demonstrates that DeepMoveSORT outperforms existing trackers in scenarios featuring non-linear motion, surpassing state-of-the-art results on three such datasets. We also perform a thorough ablation study to evaluate the contributions of different tracker components which we proposed. Based on our study, we conclude that using a learnable filter instead of the Kalman filter, along with appearance-based association is key to achieving strong general tracking performance.

\end{abstract}

% \begin{highlights}
% \item Learnable filters outperform Kalman Filter in tracking dynamically moving objects.
% \item Transformer architecture makes a good default choice for a learnable filter.
% \item Learnable filters and appearance similarity are crucial in case of non-linear motion.
% \item In case of crowded scenarios, motion-based association heuristics are vital.
% \end{highlights}

\date{May 2024}


\end{frontmatter}

\section{Introduction}

Object tracking is an essential tool in various fields, facilitating real-time video analysis, as well as the prediction of object movement and behavior. This is applicable, but not limited to fields such as autonomous driving~\citep{object_tracking_in_autunomous_driving, nuScenes, tracking_ad}, sports analysis~\citep{sportsmot, soccernet}, retail~\citep{retail_analytics}, robotics~\citep{object_tracking_in_robotics}, and surveillance~\citep{object_tracking_in_surveillence}. The tracking-by-detection paradigm, popular in multi-object tracking (MOT) tasks~\citep{mot_survey}, involves a process where objects are first detected as bounding boxes in each video frame and then associated across frames to establish objects' tracks. Association methods can be broadly divided into two primary groups: motion-based and appearance-based.\footnote{Other association method groups, such as detection confidence-based association, can optionally be used to further enhance tracking performance~\citep{hybridsort}.} Motion-based methods rely on motion models to predict the tracked object's bounding box based on its motion history, and match it with one of the detected bounding boxes. In contrast, appearance-based methods depend on the object's image features. To achieve robust accuracy across different domains, the tracking algorithm must incorporate both kinds of association methods.

At the core of the motion-based association is the motion model. Identifying an effective motion model that performs consistently across various domains presents a significant challenge. This is particularly hard in scenarios involving dynamic, non-linear motion. Currently, the state-of-the-art solutions on dataset benchmarks with non-linear motion~\citep{dancetrack, sportsmot} employ the Kalman Filter (KF) even though it is limited to using linear motion models. The KF's drawbacks are usually compensated for by a rich set of heuristics and strong appearance-based association~\citep{deep_eiou, deepocsort}. On the other hand, solutions that incorporate deep learning-based motion models lag in terms of tracking accuracy, primarily due to the lack of strong heuristics and appearance-based association~\citep{motiontrack, movesort}. 

To address the limitations of the existing methods, we build upon our previous work on end-to-end learnable filters~\citep{movesort} and engineer a tracker which combines these filters with a rich set of heuristics and appearance-based association. First we propose a new transformer-based filter architecture, TransFilter, which is both fast and accurate during training and inference. Then we enhance the filter's performance in terms of accuracy and speed by refining the measurement buffering algorithm which maintains the object's motion history, based on which the model makes its predictions. We also propose a set of heuristics which exploit the cues present in the position, shape and confidence of the detected bounding box to improve the association performance. We further improve the association by refining the standard IoU association method to account for the loss of certainty about the object's position during occlusions. By combining our end-to-end filters with appearance similarity~\citep{deepsort} and the proposed heuristics, we build {\em DeepMoveSORT}, a state-of-the-art tracker optimized for datasets featuring non-linear motion.

We demonstrate DeepMoveSORT's strong tracking performance on multiple datasets featuring non-linear motion. Most notably, we outperform the state-of-the-art tracking performance in terms of HOTA~\citep{hota} on the DanceTrack dataset~\citep{dancetrack} by $1.7\%$, on SportsMOT~\citep{sportsmot} by $1.5\%$, and SoccerNet~\citep{soccernet} by $1.3\%$. 

The detailed summary of our contributions is as follows:

\begin{itemize}
    \item In order to improve the performance of motion-based association methods, we propose a new transformer-based end-to-end filter architecture, termed TransFilter, as an alternative to existing recursive end-to-end filters---RNNFilter and NODEFilter. Compared to them, TransFilter is significantly more efficient during both training and inference, and performs better in benchmarks where a short motion history suffices.
    \item We introduce a new measurement buffering algorithm compatible with all end-to-end filters which improves over the original MoveSORT's algorithm in terms of both motion prediction accuracy and speed.
    \item We propose a way to incorporate CMC (Camera Motion Compensation) transformations to update the measurement buffer in cases where the camera motion occurs.
    \item We propose the DT-IoU (Decay Threshold IoU) motion-based association method as a direct replacement for the standard IoU association method. Unlike the standard IoU method, DT-IoU uses a minimum overlap threshold that decays during an object's occlusion to account for the decreasing certainty about the object's position.
    \item We propose the HPC (Horizontal Perspective Cues) motion-based association heuristic, which considers the object's bounding box height and vertical position to reduce the number of identity switches on crowded scenes.
    \item We introduce the ATCM (Adaptive Track Confidence Modeling) heuristic, which uses detection confidence as a metric in the association process. Following Hybrid-SORT~\citep{hybridsort}, we employ a KF to model track confidence. However, we improve the KF model with a heuristic for a more accurate measurement noise approximation. The ATCM association method also leads to a reduced number of identity switches in crowded scenes. 
    \item We combine the improved end-to-end filters and proposed heuristics with appearance similarity within the ByteTrack tracking framework~\citep{bytetrack} to create the DeepMoveSORT tracker. 
    \item We perform a detailed experimental evaluation and demonstrate that the DeepMoveSORT tracker surpasses state-of-the-art trackers on datasets with dynamic non-linear motion such as DanceTrack~\citep{dancetrack}, SportsMOT~\citep{sportsmot} and SoccerNet~\citep{soccernet}. We also perform a thorough ablation study to isolate contributions of different tracker components which we proposed.
\end{itemize}

\section{Background}
\label{sec:background}

We first provide a concise overview of the MoveSORT tracker in Section~\ref{sec:background_movesort}. The proposed DeepMoveSORT framework can be seen as an engineering extension of MoveSORT for better object tracking performance. Afterwards, in Section~\ref{sec:background_essential_compoents}, we briefly go through key components required for strong performance in modern tracking-by-detection algorithms.

\subsection{MoveSORT}
\label{sec:background_movesort}

The MoveSORT tracker~\citep{movesort} is based on the SORT~\citep{sort} tracking-by-detection framework. Accordingly, it uses an object detection model to capture the bounding boxes of objects present in the frame and performs association at each video frame. This association process involves predicting the objects' future bounding boxes and pairing them with detections in the next frame. To enhance association accuracy, MoveSORT introduced deep learning-based filters capable of motion prediction and de-noising~\citep{movesort}. These filters serve as a direct alternative to the Kalman Filter (KF) in any tracking-by-detection method, with the advantage of learning non-linear motion patterns directly from the data, without relying on domain knowledge. Employing these filters significantly improves tracker performance in datasets characterized by non-linear motion. Additionally, MoveSORT proposes a hybrid association method that calculates the cost by combining a negative IOU with the L1 distance between the predicted and the detected bounding box. The Hungarian algorithm~\citep{linear_assignment} is then used to compute the minimum-cost pairing, as standard.

\textbf{End-to-End Filters}. In the SORT approach, at each frame, the prior distribution of an object's position is obtained using the KF. This prior is combined with the measurement likelihood from the object detector, through Bayesian inference, to achieve a more accurate posterior estimation. This process is called filtering. However, measurement likelihood is not commonly available and must be approximated using domain-specific knowledge. In contrast, end-to-end filters present an alternative to Bayesian inference by allowing a deep learning-based motion model to learn to combine information from both the object detector and the motion model prediction. As this filtering process is data-driven, there is no need to approximate the measurement likelihood based on domain knowledge.

The end-to-end filter takes the objects' motion history (trajectory) before the $i$-th frame as input, and outputs a motion prediction for the $i$-th frame in the form of a Gaussian distribution $\mathcal{N}(\hat{\mu}_{i}, \hat{\Sigma}_{i})$. Once the new position of the object is observed in frame $i$, filtering is conducted by considering both the object's historical movement and the new measurement. The result of this filtering process is represented by the distribution $\mathcal{N}(\tilde{\mu}_{i}, \tilde{\Sigma}_{i})$. To train the model for accurate motion prediction and filtering, the following loss function is minimized:
\begin{equation}
    L_{e2e}(\mu_{i}, \hat{\mu}_{i}, \hat{\Sigma}_{i}, \tilde{\mu}_{i}, \tilde{\Sigma}_{i}) =
    L_{nll}(\mu_{i}, \hat{\mu}_{i}, \hat{\Sigma}_{i}) + L_{nll}(\mu_{i}, \tilde{\mu}_{i}, \tilde{\Sigma}_{i}) \label{eq:end_to_end_loss}
\end{equation}
where $\mu_{i}$ is the object's ground truth position at frame $i$, and $L_{nll}$ is the negative Gaussian log-likelihood loss. During inference, the object's motion history, defined through the measurement trajectory, is buffered and only the latest measurements are used as model inputs (e.g. the last 1 or 2 seconds). 

Two end-to-end filter architectures, RNNFilter and NODEFilter, were proposed in~\citep{movesort}. The RNNFilter model employs GRU~\citep{gru} for both motion prediction and noise filtering, whereas the NODEFilter uses NODE~\citep{node} for motion prediction and GRU for noise filtering.

\textbf{Measurement Buffering Algorithm}. The history of an object's measurements, indicative of its movement over time, contains essential features for motion prediction. The measurement buffer balances historical context by continuously adding new measurements and excluding the old ones. This balance is crucial. A history context that is too short can significantly impair motion model performance, while an excessively long history may lead to both low accuracy and slow inference. 

The algorithm is best understood through the pseudo code given in Algorithm~\ref{alg:movesort_buffer}. The hyper-parameter $T_{\text{max}}$ specifies the maximum measurement age which should be maintained by retaining only the measurements from the most recent frames. However, if only the age criterion were observed, long occlusions would result in a nearly or completely empty buffer, which could adversely affect the performance of the motion model. Therefore, a minimal buffer length (i.e.\ a lower bound on the number of measurements held in the buffer) is specified by the hyper-parameter $L_{\text{min}}$, as it can be seen in Algorithm~\ref{alg:movesort_buffer}. Properly adjusting $L_{\text{min}}$ is crucial for achieving good motion model performance~\citep{movesort}. 

\begin{algorithm}
\caption{MoveSORT's measurement buffering algorithm}
\label{alg:movesort_buffer}
\textbf{Hyper-parameters:} maximal measurement age $T_{\text{max}}$, minimal buffer length $L_{\text{min}}$

\begin{algorithmic}[1]

\Procedure{update\_buffer}{$B,x,t$}\Comment{buffer $B$, measurement $x$, time $t$}
    \If{$x$ is not null}
        \State $B \gets \text{push}(B, x)$
    \EndIf

    \If{$t - \text{time\_first}(B) \geq T_{max}\ \textbf{and}\ \text{size}(B) > L_{min}$}
        \State $B \gets \text{pop\_first}(B)$
    \EndIf
\EndProcedure
\end{algorithmic}
\end{algorithm}

\subsection{Essential Components of High-Performance Trackers}
\label{sec:background_essential_compoents}

In this section we briefly explain the main components of modern, high-performance trackers. This includes the ByteTrack~\citep{bytetrack} tracking algorithm, camera motion compensation (CMC), appearance-based association, and Hybrid-SORT's~\citep{hybridsort} track confidence modeling.

\textbf{ByteTrack}. SORT tracker~\citep{sort} uses detections whose confidence exceeds a defined threshold ({\em detection threshold} in further text). Setting this threshold to higher values reduces the number of detection false positives and, consequently, the number of incorrect tracks in the tracker but also increases the number of detection false negatives. False negatives often occur when objects are partially occluded, leading to uncertainty in the object detection model. The detection threshold value should be optimally chosen to enhance tracker performance.

ByteTrack~\citep{bytetrack} enhances the SORT approach by also considering low-confidence detections. It performs a two-step cascaded association between tracks and detections. In the first cascade, an association is made between tracks and bounding boxes with high detection scores (as defined by the confidence threshold). In the second cascade, the remaining unmatched tracks (if there are any) are associated with low-confidence detections. Unmatched detections with high confidence scores are used for the initiation of new tracks. Thus, ByteTrack prioritizes the association of visible objects, which have high detection confidence, but also performs association of partially occluded objects, which have low detection confidence. It makes tracker performance less dependant on the confidence threshold hyper-parameter and improves the tracker performance on multiple datasets~\citep{bytetrack}.

\textbf{Camera Motion Compensation (CMC)}. Deep learning-based motion models can learn to adapt to camera motion when the motion follows predictable patterns. This adaptation is particularly useful for camera movements during sports games, which typically shift smoothly from left to right or vice versa. However, camera motion is not always predictable, for instance, when a person holds a camera phone loosely or during sudden camera shifts caused by wind. In these scenarios, estimating camera motion from video image features is more effective. 

CMC techniques based on image features are commonly employed in the tracking-by-detection paradigm to counteract the negative effects of camera movement during tracking. These techniques usually involve a comparison of the extracted image features of the current and previous frames to approximate the camera motion. The result is a transformation matrix from the coordinate system of the previous frame to the coordinate system of the current frame. This transformation matrix allows us to apply corrections to the predicted bounding box coordinates.

\textbf{Appearance-based association}. Appearance-based association enhances tracking by leveraging deep learning to extract object appearance features which are used for object re-identification across frames. It presents a key component of modern tracking-by-detection algorithms~\citep{botsort, bytetrack, deepocsort, hybridsort}. DeepSORT~\citep{deepsort}, the foundation of modern trackers, incorporates appearance-based association into the SORT framework and combines it by simple cost weighting with motion-based association to boost re-identifaction accuracy. 

\textbf{Hybrid-SORT's Track Confidence Modeling}. In crowded scenes with significant overlap among multiple objects, differentiation based on detection confidence scores becomes useful. Objects at the forefront, which are less occluded, typically receive higher confidence scores from the detector, whereas those in the background are assigned lower scores. Motivated by this observation, Hybrid-SORT~\citep{hybridsort} incorporates the object detection confidence values into the association process. It implements a Track Confidence Modeling (TCM) method that employs a Kalman filter to predict a tracked object's detection confidence. The difference between the predicted confidence and that produced by the object detector is used as another cue for associating tracks with detected bounding boxes. This method can be easily integrated with other association criteria by simple weighting.

\section{Methodology}
\label{sec:methodology}

In this section, we present the main methodological contributions of our work, focusing on object trackers that incorporate deep learning-based motion models and filters. First, we define the optimization objective for object trajectory prediction and filtering tasks. Afterwards, we introduce a new end-to-end architecture---TransFilter. Furthermore, we describe the improvements to the measurement buffering algorithm, which enhance the performance of all end-to-end filters. Subsequently, we introduce a new set of heuristics to improve tracking association in crowded scenarios. Finally, we construct a high-performance tracker by combining deep learning-based filters, appearance features, and the proposed heuristics.

\subsection{Object trajectory prediction and filtering}

We define the tasks of trajectory prediction and filtering (i.e.\ denoising) analogously to how they were defined for the end-to-end filters~\citep{movesort}. The input trajectory is observed at time points $\bm{T}_{O} = \{t_1, t_2, t_3, \dots , t_n\}$ with corresponding observation vector values $\bm{X}_{O} = \{x_1, x_2, x_3, $\dots$ x_n\}$. These time points do not need to be equidistant. The \emph{prediction task} involves forecasting the target trajectory $\bm{X}_{T} = \{x_{n+1}, x_{n+2}, x_{n+3}, \dots, x_{n+m}\}$ at target time points $\bm{T}_{T} = \{t_{n+1}, t_{n+2}, \dots, t_{n+m}\}$. The \emph{filtering task} refers to the incorporation of new measurements, observed at times $T_T$, so that the noise from these observation is suppressed and the predicted trajectory $X_T$ is improved. An end-to-end filter model is optimized for both tasks simultaneously.

Strong tracker performance does not necessitate uncertainty estimation for motion prediction as the prediction uncertainty is usually not used during track association~\citep{sort, bytetrack, botsort, sparsetrack, imprasso}. In addition, end-to-end filters do not require uncertainty estimation for noise filtering either~\citep{movesort}. For these reasons, we simplify the task to prediction (and filtering) of the trajectory, and we do not estimate the parameters of a Gaussian distribution for each time point of the trajectory, in contrast to the approach described in the background Section~\ref{sec:background_movesort}. This approach not only simplifies the optimization problem but also offers greater flexibility in selecting the loss functions.

We further redefine the loss function $L_{e2e}$ for non-probabilistic end-to-end models as follows:
\begin{equation}
    L_{e2e}(\mu_{i}, \hat{\mu}_{i}, \tilde{\mu}_{i}) = L_{predict}(\mu_{i}, \hat{\mu}_{i}) + L_{update}(\mu_{i}, \tilde{\mu}_{i}) \label{eq:end_to_end_loss}
\end{equation}
where $\mu_{i}$ represents the ground truth value, $\hat{\mu}_{i}$ the model-predicted value and $\tilde{\mu}_{i}$ the updated value after the $i$-th measurement is observed. For both $L_{predict}$ and $L_{update}$, we employ the Huber loss~\citep{huber} with the hyper-parameter $\delta$, as it offers greater robustness to outliers compared to the mean squared error loss.

\textbf{Feature extraction}. Feature extraction is performed on both the input and target trajectories to obtain richer features. Extracted features include absolute coordinates, coordinates relative to the last observation and coordinate differences between consecutive time points. Using these features instead of just the absolute bounding box coordinates is crucial for strong motion model performance~\citep{movesort}. These transformed trajectories with rich features are then employed as inputs to the model. Transformations are independent of the model architecture. Details regarding the trajectory feature extraction can be found in \ref{appendix:trajectory_features}.

In the following section, we introduce a novel end-to-end filter architecture trained using the proposed training procedure.

\subsection{TransFilter}

RNNFilter and NODEFilter, which are architectures within the end-to-end deep learning-based filter family~\citep{movesort}, offer strong motion prediction accuracy boost in datasets characterized by non-linear motion. Additionally, they require no hyper-parameter tuning for specific object detection models after model training. However, their recursive approach to encoding trajectory features and motion prediction results in slower inference\footnote{Recursive feature encoding can be efficient if the features of old measurements remain unchanged upon the observation of a new measurement. This is not the case for our trajectory features.} and longer training times\footnote{NODE architectures are well-known for slower inference and, particularly, extended training durations~\citep{node, movesort}.}. Additionally, they are prone to error accumulation during motion prediction due to their recursive state updates~\citep{movesort}.

\begin{figure*}%[H]
  \centering
  \includegraphics[width=0.5\linewidth]{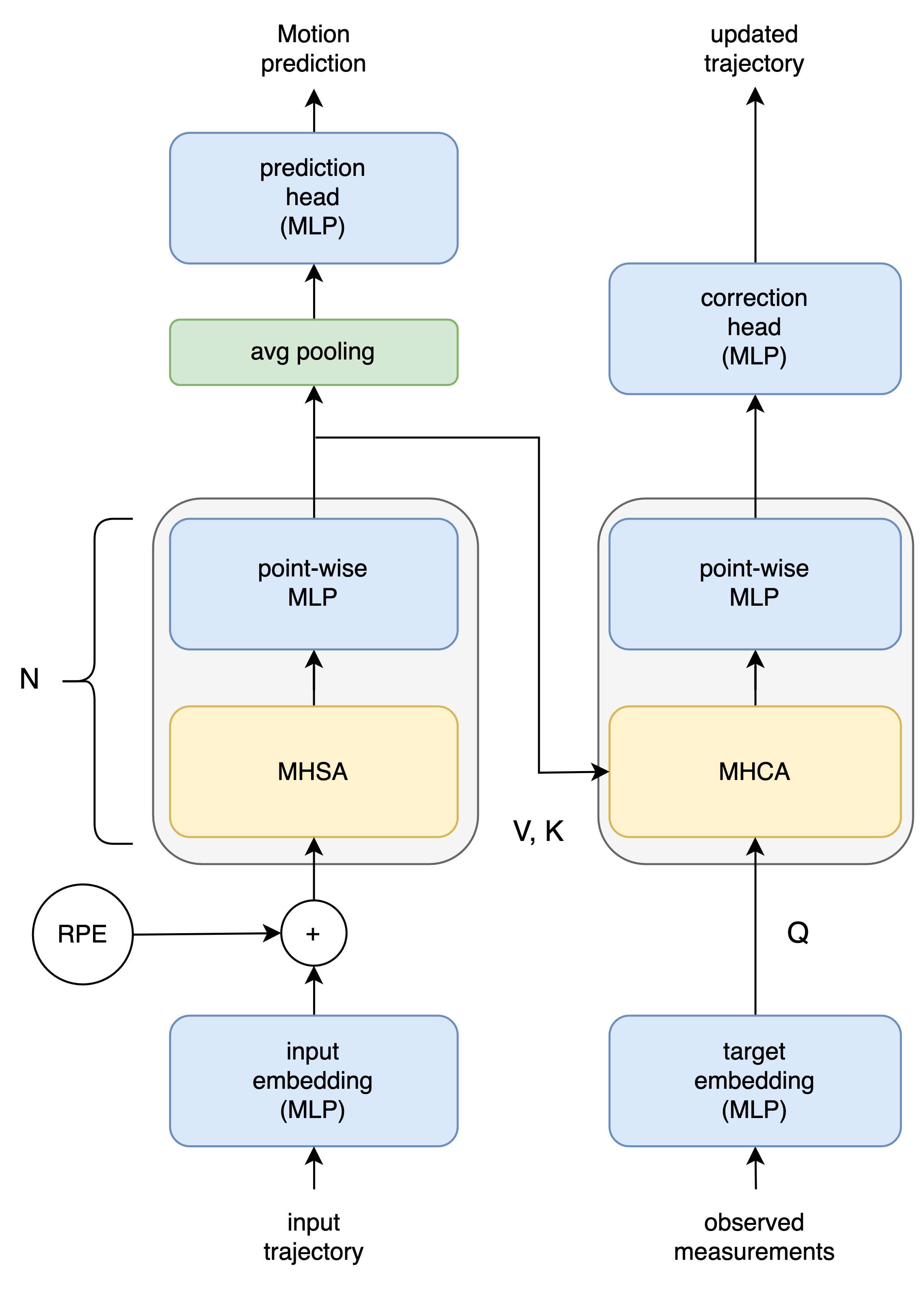}
  \caption{Overview of the TransFilter architecture for motion prediction (encoder) and noise filtering (decoder). The input trajectory is first transformed into token embeddings. Combined with RPE positional encoding, these token embeddings serve as input to the stacked transformer encoder consisting of $N$ layers. Motion prediction is performed based on the average-pooled encoder outputs with one-shot multi-step prediction. The encoder outputs are also used as inputs to the decoder, combined with the observed measurements. The decoder performs corrections for each measurement input.}
  \label{fig:transfilter}
\end{figure*}

As an alternative to the previously mentioned architectures, we propose TransFilter, a transformer-based end-to-end filter architecture. It can perform feature encoding and motion prediction for multiple steps in parallel, making it efficient during training and inference. We extend the standard transformer architecture~\citep{attention_is_all_you_need} to an end-to-end filter. Our architecture, shown in Figure~\ref{fig:transfilter}, consists of two core components: encoder for motion prediction and decoder for noise filtering.

\textbf{Token embedding}. We employ a Multi-Layer Perceptron (MLP) to obtain token embeddings for each measurement within the input trajectory. These tokens, summed with positional encoding, serve as inputs to the transformer encoder. As an alternative to the conventional positional encoding described in~\citep{attention_is_all_you_need}, we use Reversed Positional Encoding (RPE)~\citep{rpe}. RPE formulation was initially proposed for right-to-left text, such as Arabic dialects; however, it is not semantically equivalent when applied to time-series. In contrast to the (left-to-right) natural language processing in which the first word should always receive the same recognizable positional encoding, in our case the last observation is of central importance, since it is the closest to the step being predicted. Therefore the last observation should always receive the same positional encoding. Hence, our approach applies positional encoding starting from the end. This process is equivalent to the next three steps: reversing the sequence, applying standard positional encoding, and then reversing the sequence again. So, rather than indicating the position of the measurement in the trajectory relative to the beginning, our positional encoding indicates how recent (close to the end) the measurement is.

\textbf{Motion Prediction}. After acquiring the token embeddings, we use them as inputs for the transformer encoder. We apply average pooling on the final encoder outputs over the temporal dimension. The aggregated output from the encoder is then used for single-shot multi-step prediction to generate all required predictions in one go through the motion prediction head, as illustrated in Figure~\ref{fig:transfilter}. The primary advantage of employing single-shot prediction, as opposed to the recursive approach used in the aforementioned end-to-end filters, is the elimination of error accumulation. In this case, predictions for all target time points are made simultaneously. However, this strategy might result in less natural and less smooth trajectory prediction.

\textbf{Noise filtering}. During training, the decoder takes token embeddings of observations corresponding to the target trajectory as input. We consider the outputs from the decoder to be the corrected measurements with suppressed noise. During inference, we perform filtering as the object's measurements are observed, i.e.\ one at a time. This makes the multi-head self-attention in the decoder redundant during inference, so we remove it from the architecture, as illustrated in Figure~\ref{fig:transfilter}. 

Compared to the RNNFilter and NODEFilter architectures, which interleave prediction and filtering during training, the TransFilter architecture completely decouples the components for these tasks. This leaves an option to omit the decoder and employ solely the encoder for the motion model, in cases when the object detector is sufficiently reliable so that the filtering is unnecessary once the correct observation is associated with the track.

\subsection{Improved filter's measurement buffer}
\label{sec:improved_measurement_buffer}

This section introduces two modifications that enhance the filter's measurement buffer, a core component shared across all deep learning-based filter architectures we use in this paper. Consequently, these improvements yield benefits for the entire family of these architectures. First we modify the old measurement buffering algorithm and then we define the process of aligning the buffer with camera motion.

\textbf{Improved measurement buffering algorithm}. We have refined the measurement buffering algorithm described in Section \ref{sec:background_movesort}, to enhance its effectiveness during occlusions. The improved algorithm is specified by the pseudo code in Algorithm~\ref{alg:deepmovesort_buffer}. We modify the original algorithm behaviour during occlusions in two respects. First, we preserve all measurements in the buffer until a new measurement is detected. This diverges from the prior approach of removing the oldest measurement at every frame\footnote{The oldest measurement is removed only in~\citep{movesort} if it was observed at that frame i.e.\ if the object was detected in that frame.} unless the history must be preserved due to a low number of measurements in the buffer. As a result, our buffer retains an adequate and consistent number of measurements throughout the long occlusion so that motion predictions can be made. Consequently, the criteria and the hyper-parameter for history preservation have been rendered unnecessary. Second, upon the detection of a new measurement after the object's occlusion, outdated measurements are systematically excluded from the buffer, as seen from the \textit{while} loop in Algorithm~\ref{alg:deepmovesort_buffer}. This strategy favors a shorter, more recent measurement history over a protracted, fragmented one, with measurements that are already outdated once the occlusion ends. This modification leads to a more efficient measurement buffering algorithm when tracking dynamic objects. 

Another significant benefit of our proposed measurement buffering algorithm is the improved inference speed during occlusions. Since the inputs do not change during an occlusion, there is no need to rerun the encoder at each step. Instead, encoder outputs can be saved for occluded objects in order to reduce the computational cost.

\begin{algorithm}
\caption{DeepMoveSORT's measurement buffering algorithm}
\label{alg:deepmovesort_buffer}
\textbf{Hyper-parameters:} $T_{\text{max}}$

\begin{algorithmic}[1]

\Procedure{update\_buffer}{$B,x,t$}\Comment{buffer $B$, measurement $x$, time $t$}
    \If{$x$ is not null}
        \State $B \gets \text{push}(B, x)$
        \While{$t - \text{time\_first}(B) \geq T_{\text{max}}$}
            \State $B \gets \text{pop\_first}(B)$
        \EndWhile
    \EndIf
\EndProcedure
\end{algorithmic}
\end{algorithm}

\textbf{Aligning buffered measurements to camera motion}. We assume that a CMC method is used to obtain an affine transformation matrix $A^{(i)}$ from the coordinate system of frame $i-1$ to the coordinate system of frame $i$. 

The acquired affine transformation matrix is used to map buffered trajectory measurements, i.e.\ points, between the frame coordinate systems. We recursively define the affine transformation from frame $i-1$ to frame $i $ of the measurement $x_k$ observed at frame $k$ where $i > k$ as follows:
\begin{equation}
    x^{(i)}_{k} = A^{(i)} x^{(i-1)}_{k} = A^{(i)} A^{(i-1)} \ldots A^{(k+1)} x_{k}\label{eq:measurement_cmc} 
\end{equation}
This is equivalent to applying the transformation to all measurements in the buffer at each frame. This extension of the measurement buffering algorithm is important for a high-performance tracker in scenarios where camera motion is unpredictable.

\subsection{Association method}
\label{sec:association_methods}

In this section, we first introduce a novel association method as a replacement for both the standard Intersection over Union (IoU) association cost and the gating mechanism. Subsequently, we introduce improved variants of the existing heuristics to further enhance tracking performance. Lastly, we define the process of obtaining and using the objects' appearance features in conjunction with these heuristics.

\textbf{Decay Threshold IoU}. The SORT~\citep{sort} tracking method prevents matching tracks with objects detected far away from each other by employing the IoU gating. A track cannot be matched with an object if their IoU is below a predefined threshold. This minimum IoU threshold is constant for all tracks, regardless of whether they are active or occluded. However, this approach does not account for the decreased certainty in the position of an object, the longer it remains occluded. Ideally, the threshold should be higher for active tracks and lower for occluded objects to accommodate for this uncertainty. This adjustment aligns with the observation that during an occlusion the accuracy of the motion model decreases with each subsequent step forward.

We introduce the Decay Threshold IoU (DT-IoU) association method, which is based on a heuristic that employs a variable minimum IoU threshold depending on the time the object has been occluded. The proposed association threshold $IoU_{\text{min}}$ is defined as a function of the object's occluded time $t_{\text{occluded}}$ as follows:
\begin{equation}
IoU_{\text{min}}(t_{\text{occluded}}) = \max(IoU_{\text{upper\_bound}} - IoU_{\text{decay}} \cdot t_{\text{occluded}}, IoU_{\text{lower\_bound}})
\label{eq:dt_iou} 
\end{equation}
where $IoU_{\text{upper\_bound}}$ is the starting threshold used for active tracks, $IoU_{\text{decay}}$ is the decay rate, and $T_{\text{lower\_bound}}$ is the lowest possible value for the threshold.

When tracking objects with very fast and dynamic movements, the overlap between the predicted and detected bounding boxes can be very small or non-existent. A direct method to compensate for these scenarios is to perform an expansion of the bounding boxes before calculating the IoU~\citep{deep_eiou, cbiou}. In scenarios where it proves beneficial (e.g., tracking sports players), we adjust the scales of the bounding boxes before computing the IoU cost. This adjustment involves enlarging the bounding boxes horizontally and vertically from the center, by employing an expansion rate denoted by $E_{rate}$.

\textbf{Horizontal Perspective Cues}. The hybrid association method introduced in MoveSORT~\citep{movesort} combines IoU and L1 costs between the extrapolated tracks and detected objects to improve association accuracy in crowded scenes with frontal-view cameras. Unlike IoU, the L1 distance takes into account the scale of objects, which diminishes for objects that are further away. Furthermore, due to perspective, objects that are further away appear higher on the y-axis on the coordinate system of the image frame. Instead of simply using the L1 distance, we focus on the bounding box's height and bottom position, as these tend to fluctuate less over time compared to the width and x-axis position~\citep{hybridsort, sparsetrack} and hence their change is more relevant. We define the Horizontal Perspective Cues (HPC) cost $C_{HPC}$ between the predicted track bounding box $\hat{x}$ and the detection $D$ as:
\begin{equation}
C_{HPC}(\hat{x}, D) = \lambda_{h} \cdot |\hat{x}_h - D_h| + \lambda_{y} \cdot |\hat{x}_y - D_y|
\label{eq:hpc} 
\end{equation}
where $\hat{x}_h$ and $D_h$ are the heights of the predicted bounding box and detection, respectively, $\hat{x}_y$ and $D_y$ are their bottom y-axis positions, and $\lambda_{h}$ and $\lambda_{y}$ are the cost weights.

\textbf{Adaptive Track Confidence Modeling (ATCM)}. We apply a simple modification to the detection confidence modeled by the KF in the Hybrid-SORT's TCM heuristic~\citep{hybridsort}. Instead of using a constant measurement noise variance, we define it to be dependant on the detected object confidence as $(\sigma_{conf} \cdot (1 - D_{conf}))^2$, where $\sigma_{conf}$ is the constant standard deviation multiplier and $D_{conf}$ is the detection confidence. As noted in Hybrid-SORT~\citep{hybridsort}, the high confidence detections are less noisy and less prone to dynamic changes. This allows us to apply this knowledge to the KF. 

\textbf{Appearance modeling}. For appearance features, we employ ReID (re-identification) model from the FastReID framework~\citep{fastreid}, in line with the methodologies of previous works~\citep{botsort, hybridsort}. Upon detecting objects in a frame, we crop the object patches, pre-process them, and feed them into the ReID model to extract appearance features. For associating the track's and object's appearance features, we employ cosine distance, calculated for each (track, detection) pair. The object track's appearance features are computed using an exponential moving average (EMA) of the object's appearance features from previous frames. Appearance association is used only in the first ByteTrack's association cascade, in which the detections have a high confidence score meaning that we expect useful appearance features. 

\textbf{Association}. We combine the costs of the DT-IoU association method with HPC, and ATCM heuristics with the cosine distance between the appearance embeddings into a single association cost matrix, which is used in the Hungarian algorithm. We define this matrix as follows:
\begin{equation}
C = \lambda_{DT-IoU} \cdot C_{DT-IoU} + \lambda_{HPC} \cdot C_{HPC} + \lambda_{ATCM} \cdot C_{ATCM} + \lambda_{Appr} \cdot C_{Appr}
\label{eq:hpc} 
\end{equation}
where $C_{DT-IoU}$, $C_{HPC}$, $C_{ATCM}$ and $C_{Appr}$ are costs with corresponding weights $\lambda_{DT-IoU}$, $\lambda_{HPC}$, $\lambda_{ATCM}$ and $\lambda_{Appr}$. Weighted cost matrix is used as input to the Hungarian algorithm to obtain optimal matches between the tracks and detected objects in the current frame. These weights should be balanced based on the relative importance of motion versus appearance.

\subsection{DeepMoveSORT}

Our proposed DeepMoveSORT method enhances MoveSORT in five main aspects:
\begin{itemize}
    \item It uses a more efficient ByteTrack baseline, in comparison to the SORT baseline.
    \item It enhances the accuracy of all motion models with improved buffering methods, thereby improving motion-based association.
    \item It uses CMC for scenes with non-predictable camera motion.
    \item It integrates motion-based association with appearance-based association.
    \item It employs DT-IoU, HPC and ATCM heuristics to achieve better association performance in crowded scenes where occlusions frequently occur.
\end{itemize}

We refer to the tracker encompassing all proposed improvements as the DeepMoveSORT tracker. 

\section{Experimental evaluation}
\label{sec:experiments}

We evaluate the proposed methods using five MOT datasets---DanceTrack~\citep{dancetrack}, SportsMOT~\citep{sportsmot}, SoccerNet~\citep{soccernet}, MOT17, and MOT20~\citep{mot20}. Our evaluation relies on the higher order metrics (HOTA, AssA, DetA), IDF1, and MOTA---Multi-Object Tracking Accuracy~\citep{hota}. The exact set of used metrics depends on the dataset.

In subsequent sections, we first define our experimental setup, then evaluate and analyze the results on the mentioned MOT datasets. Lastly, we conduct an experimental ablation study and discussion.

\subsection{Technical details}
\label{sec:experimental_setup}

In this section, we briefly discuss design choices for object detection, ReID, CMC, and filters. Tracker configuration (hyper-parameters) for each dataset can be found in~\ref{appendix:tracker_hyperparameters}.

\textbf{Object detection}. We use the publicly available YOLOX~\citep{yolox} models for evaluation on all dataset benchmarks. For ablation study on the SportsMOT validation set we train a YOLOv8~\citep{yolov8} instead of YOLOX, due to hardware limitations. Detailed model configurations for each dataset can be found in~\ref{appendix:object_detection}.

\textbf{ReID}. For appearance feature extraction, we employ BoT-S50 (Bag of Tricks)~\citep{bot} ReID models from the FastReID framework~\citep{fastreid}. Detailed information about each ReID model for each dataset can be found in~\ref{appendix:reid}.

\textbf{CMC}. For the CMC algorithm, we follow~\citep{botsort} and employ GMC video stabilization module from the OpenCV library~\citep{opencv_library}. We use CMC only on the MOT17 dataset, since it is the only dataset with frequent unpredictable camera motion. We did not see significant improvements (i.e, more than 0.1 as measured in terms of HOTA) when we applied CMC to MOT20, DanceTrack, SportsMOT, or SoccerNet datasets.

\textbf{Filters}. We train the TransFilter and the {\em improved RNNFilter} -- RNNFilter employing our proposed measurement buffering algorithm. We do not consider the NODEFilter, as it is similar to the RNNFilter in terms of precision, but is slower during training and inference. We use the Bot-Sort's implementation of the KF~\citep{botsort} for our experimental analysis. Details of the filter training setup can be found in~\ref{appendix:training_filters}.

\subsection{Benchmark results}
\label{sec:benchmark_results}

This section includes benchmark results for DanceTrack, SportsMOT, SoccerNet, MOT17 and MOT20 datasets. 

\textbf{DanceTrack}. DanceTrack is a dataset comprising 100 videos of people dancing, each meticulously annotated for the purpose of multi-object tracking of dancers. This task is particularly challenging due to the dancers' similar appearances, frequent occlusions, and their rapid, non-linear movements. The dataset is segmented into 40 training videos, 25 validation videos, and 35 test videos~\citep{dancetrack}.

We compare the performance of our tracker to other published methods on the DanceTrack test set. The results are shown in Table~\ref{tab:results_test_dancetrack}. All trackers presented use the same DanceTrack's public object detection model. Our DeepMoveSORT-TransFilter outperforms state-of-the-art Deep OC\_SORT by $1.7\%$ in HOTA, $0.8\%$ in MOTA and $3.5\%$ in IDF1, showing very strong performance. Additionally, our method improves upon MoveSORT by $6.9\%$ in HOTA, $0.8\%$ in MOTA and $9.0\%$ in IDF1. We can also observe that DeepMoveSORT-TransFilter outperforms the DeepMoveSORT-RNNFilter variant by $3.6\%$ in HOTA, $0.1\%$ in MOTA and $4.0\%$ in IDF1, which demonstrates the TransFilter model's benefits compared to the improved RNNFilter in crowded scenes.

For a fair comparison, we separate UCMCTrack~\citep{ucmc} and Hybrid-SORT~\citep{hybridsort} as the information from the test set were used for these methods. UCMCTrack requires camera parameter tuning in advance on each scene (including test scenes), while Hybrid-SORT uses different hyper-parameters on validation and test sets as can be seen from its GitHub repository.\footnote{\url{https://github.com/ymzis69/HybridSORT}, commit SHA: 396f8d30db13304c0cbaf1dcf2e16ded93ce1701} 

\begin{table*}
\small
\centering
\begin{tabular}{l|ccccc}
Method & HOTA & DetA & AssA & MOTA & IDF1 \\
\hline
\textit{w/ tuning on test set} & \multicolumn{5}{c}{} \\
UCMCTrack~\citep{ucmc}& 63.6 & - & 51.3 & 88.9 & 65.0 \\
Hybrid-SORT~\citep{hybridsort} & 65.7 & - & - & 91.8 & 67.4 \\
\hline\hline
\textit{w/o tuning on test set} & \multicolumn{5}{c}{} \\
DeepSORT~\citep{deepsort} & 45.6 & 71.0 & 29.7 & 87.8 & 47.9 \\
ByteTrack~\citep{bytetrack} & 47.3 & 71.6 & 31.4 & 89.5 & 52.5 \\
SORT~\citep{sort} & 51.2 & 80.5 & 32.6 & 91.5 & 49.7 \\ 
MotionTrack (DanceTrack)~\citep{motiontrack} & 52.9 & 80.9 & 34.7 & 91.3 & 53.8 \\
BoT-SORT~\citep{botsort} & 53.6 & 79.9 & 36.1 & 92.3 & 55.3 \\
BoT-SORT-ReID~\citep{botsort} & 54.2 & 80.0 & 36.8 & 92.1 & 57.0 \\
OC\_SORT~\citep{ocsort} & 55.1 & 80.4 & 38.0 & 89.4 & 54.9 \\
SparseTrack~\citep{sparsetrack} & 55.5 & 78.9 & 39.1 & 91.3 & 58.3 \\
MoveSORT~\citep{movesort} & 56.1 & 81.6 & 38.7 & 91.8 & 56.0 \\
MotionTrack~\citep{motiontrack} & 58.2 & 81.4 & 41.7 & 91.3 & 58.6 \\
Deep OC\_SORT~\citep{deepocsort} & \textit{61.3} & 81.6 & \textit{45.8} & 91.8 & \textit{61.5} \\
\hline
DeepMoveSORT-TransFilter (ours) & \textbf{63.0} & \textit{82.0} & \textbf{48.6} & \textbf{92.6} & \textbf{65.0} \\
DeepMoveSORT-RNNFilter (ours) & 59.4 & \textbf{82.1} & 43.1 & \textit{92.5} & 61.0 \\
\end{tabular}
\caption{Evaluation results on the DanceTrack test set using the public DanceTrack object detection model. The best results are in bold for each metric, while the second-best results are italicized. The results for the SORT are taken from~\citep{movesort}, ByteTrack and DeepSORT from~\citep{dancetrack}, while the BoT-SORT and BoT-SORT-ReID results are produced by us.}
\label{tab:results_test_dancetrack}
\end{table*}

\textbf{SportsMOT}. SportsMOT is a comprehensive multi-object tracking dataset that includes $240$ video sequences, totaling over $150,000$ frames. The dataset encompasses three categories of sports: basketball, volleyball, and football. It is particularly challenging for object tracking due to the fast and varying speeds of motion and the similar appearances of the objects involved. The dataset is organized into training, validation, and test sets, which contain $45$, $45$, and $150$ video sequences, respectively~\citep{sportsmot}.

We compare the performance of our tracker to other published methods on the SportsMOT test set. The results are shown in Table~\ref{tab:results_test_sportsmot}. We can observe that DeepMoveSORT outperforms the state-of-the-art Deep-EIoU tracker when employing either the TransFilter or the improved RNNFilter. DeepMoveSORT-TransFilter outperforms Deep-EIoU by a small margin of $0.4\%$ in HOTA, $0.2\%$ in MOTA and $0.3\%$ in IDF1, while the DeepMoveSORT-RNNFilter outperforms it by a large margin of $1.5\%$ in HOTA, $0.2\%$ in MOTA and $1.9\%$ in IDF1. The RNNFilter works well with a longer measurement buffer, compared to TransFilter, making it robust for tracking on sport datasets. Compared to the MoveSORT~\citep{movesort} using the original RNNFilter, the DeepMoveSORT-RNNFilter outperforms it by $4.1\%$ in HOTA and $4.8\%$ in IDF1, but underperforms by $0.2\%$ in MOTA.

\begin{table*}%[H]
\centering
\small
\begin{tabular}{l|ccccc}
Method & HOTA & DetA & AssA & MOTA & IDF1 \\
\hline
FairMOT~\citep{fairmot} & 49.3 & 70.2 & 34.7 & 86.4 & 53.5 \\
QDTrack~\citep{qdtrack} & 60.4 & 77.5 & 47.2 & 90.1 & 62.3 \\
CenterTrack~\citep{centertrack} & 62.7 & 82.1 & 48.0 & 90.8 & 60.0 \\
ByteTrack~\citep{bytetrack} & 64.1 & 78.5 & 52.3 & 95.9 & 71.4 \\
MixSORT-Byte~\citep{sportsmot} & 65.7 & 78.8 & 54.8 & 96.2 & 74.1 \\ 
BoT-SORT~\citep{botsort} & 68.7 & 84.4 & 55.9 & 94.5 & 70.0 \\
TransTrack~\citep{transtrack} & 68.9 & 82.7 & 57.5 & 92.6 & 71.5 \\
OC\_SORT~\citep{ocsort} & 73.7 & 88.5 & 61.5 & 96.5 & 74.0 \\
MotionTrack~\citep{motiontrack} & 74.0 & 88.8 & 61.7 & 96.6 & 74.0 \\
MixSORT-OC~\citep{sportsmot} & 74.1 & 88.5 & 62.0 & 96.5 & 74.4 \\ 
MoveSORT~\citep{movesort} & 74.6 & 87.5 & 63.7 & \textit{96.7} & 76.9 \\
Deep-EIoU* ~\citep{deep_eiou} & 77.2 & \textbf{88.2} & 67.7 & 96.3 & 79.8 \\
\hline
DeepMoveSORT-TransFilter* (ours) & \textit{77.6} & \textbf{88.2} & \textit{68.3} & 96.5 & \textit{80.1} \\
DeepMoveSORT-RNNFilter* (ours) & \textbf{78.7} & \textit{88.1} & \textbf{70.3} & 96.5 & \textbf{81.7}
\end{tabular}
\caption{Evaluation results on the SportsMOT test. The best results are in bold for each metric, while the second-best results are italicized. The results for the BoT-SORT are taken from~\citep{deep_eiou}. Trackers marked with $*$ share the same object detector. }
\label{tab:results_test_sportsmot}
\end{table*}

\textbf{SoccerNet}. SoccerNet is an extensive dataset featuring interesting moments from 12 different soccer games. Key moments from these games are selected and presented as 30-second clips, which are then annotated. The dataset is organized into three sets: training, test, and public challenge, containing $57$, $49$, and $95$ clips, respectively. Due to the similarities of the domain (soccer), we use the same tracker configuration as in SportsMOT, including the object detection, ReID, filter models and tracker hyper-parameters.

We compare the performance of our tracker against other published methods on the SoccerNet test set. As shown in Table~\ref{tab:results_test_soccernet}, our DeepMoveSORT-RNNFilter outperforms all competing methods, including BoT-SORT-ReID, with a $2.0\%$ improvement in HOTA and a $4.0\%$ improvement in AssA. However, it underperforms by $0.1\%$ in DetA. Furthermore, DeepMoveSORT-RNNFilter surpasses DeepMoveSORT-TransFilter by $0.3\%$ in HOTA and $0.7\%$ in AssA but falls behind by $0.2\%$ in DetA. This is expected as the SoccerNet shares the same domain with the SportsMOT where the improved RNNFilter also performs well.

\begin{table*}%[H]
\centering
\small
\begin{tabular}{l|ccc}
Method & HOTA & DetA & AssA \\
\hline
DeepSORT~\citep{deepocsort} & 36.6 & 40.0 & 33.8 \\
FairMOT~\citep{fairmot} & 57.9 & \textbf{66.6} & 50.5 \\
ByteTrack & 59.8 & 65.6 & 54.7 \\
ByteTrack-ReID & 61.5 & 65.8 & 57.6 \\
BoT-SORT & 59.8 & 65.6 & 54.7 \\
BoT-SORT-ReID & 60.8 & 65.7 & 56.2 \\
\hline
DeepMoveSORT-TransFilter (ours) & \textit{62.5} & \textit{65.8} & \textit{59.5} \\
DeepMoveSORT-RNNFilter (ours) & \textbf{62.8} & 65.6 & \textbf{60.2}
\end{tabular}
\caption{Evaluation results on the SoccerNet test set. The best results are in bold for each metric, while the second-best results are italicized. The results for the DeepSORT and FairMOT are taken from~\citep{soccernet}, while the results for ByteTrack, ByteTrack-ReID, BoT-SORT and BoT-SORT-ReID are produced by us.}
\label{tab:results_test_soccernet}
\end{table*}

Evaluation on SoccerNet with oracle detections\footnote{We define oracle detections as ground truth bounding boxes that are used in place of actual detections during tracking inference.} can be found in~\ref{appendix:additional_benchmarks}.

\textbf{MOTChallenge---MOT17 and MOT20} The MOT17 dataset includes 14 short video scenes with non-static cameras and predominantly linear pedestrian motion, split equally between training and test sets. Similarly, the MOT20 dataset contains eight crowded scenes with static cameras, also evenly divided between training and test sets, but with approximately three times more bounding boxes in total compared to MOT17~\citep{mot20}. The predominately linear motion in these scenes means that the motion models we propose are {\em not expected} to provide performance benefits. Instead, we include them only for completeness, since they are standard benchmarks in the field.

In accordance with the MOTChallenge guidelines,\footnote{MOTChallenge (\url{https://motchallenge.net}) is a platform dedicated to the evaluation of multiple object tracking algorithms. It provides standardized datasets with ground truth annotations, enabling consistent and equitable comparisons among various object tracking approaches.} we submit only one method to the test server. Following the established protocol, we split each training scene into two parts: the first half is used for the training set, and the second half is used for the validation set~\citep{bytetrack, botsort, deepocsort, sparsetrack, motiontrack_byte_cmc}. We evaluate the TransFilter, improved RNNFilter, and KF on the MOT17 and MOT20 validation sets, with results for both datasets shown in Table~\ref{tab:results_val_mot17_and_mot20}. In this case, the TransFilter is outperformed by the KF. We submit DeepMoveSORT-TransFilter for evaluation on both MOT17 and MOT20 test sets, as it has noticeably better performance than the RNNFilter. 

\begin{table*}%[H]
\centering
\small
\begin{tabular}{l|ccc|ccc}
& \multicolumn{3}{c|}{MOT17} & \multicolumn{3}{c}{MOT20} \\
Method & HOTA & IDF1 & MOTA & HOTA & IDF1 & MOTA \\
\hline
DeepMoveSORT-KF & \textbf{70.3} & \textbf{82.9} & \textbf{80.3} & \textbf{59.0} & \textbf{74.6} & \textbf{77.1} \\
DeepMoveSORT-TransFilter & 70.1 & 82.8 & 80.1 & \textbf{59.0} & 74.5 & \textbf{77.1} \\
DeepMoveSORT-RNNFilter & 69.6 & 81.9 & 79.9 & 58.0 & 72.9 & 76.7
\end{tabular}
\caption{Comparison of our trackers on the MOT17 and MOT20 validation sets.}
\label{tab:results_val_mot17_and_mot20}
\end{table*}

The results of our method on the MOT17 and MOT20 benchmarks are presented in Table~\ref{tab:results_test_mot17_and_mot20}. We observe that our method underperforms compared to state-of-the-art methods on MOT17 and MOT20. This outcome is anticipated, as our method has the high flexibility required for learning non-linear motion patterns, which makes it prone to overfitting on simpler datasets. Specifically, on the MOT17 test set, our method is outperformed by SUSHI~\citep{pia2} by of $3.3\%$ in HOTA, $2.4\%$ in IDF1, and $4.4\%$ in MOTA. On the MOT20 test set, our method is outperformed by PIA2 by $4.1\%$ in HOTA, $5.9\%$ in IDF1, and $4.9\%$ in MOTA. However, it is noteworthy that the performance gap between our method and some state-of-the-art methods on datasets featuring non-linear motion is considerably larger; for example, our method outperforms ByteTrack by $9.7\%$ in HOTA on DanceTrack and by $10\%$ in HOTA on SportsMOT. Compared to the other benchmarks, MOTChallenge has almost no benefits from deep learning-based motion models and appearance-based association.

\begin{table*}%[H]
\centering
\small
\begin{tabular}{l|ccc|ccc}
& \multicolumn{3}{c|}{MOT17} & \multicolumn{3}{c}{MOT20} \\
Method & HOTA & MOTA & IDF1 & HOTA & MOTA & IDF1 \\
\hline
MotionTrack~\citep{motiontrack} & 60.9 & 76.5 & 73.5 & 58.3 & \textit{75.0} & 53.9 \\
MoveSORT~\citep{movesort} & 62.2 & 79.5 & 75.1 & 60.5 & 74.3 & 74.0 \\
ByteTrack~\citep{bytetrack} & 63.1  & 80.3 & 77.3 & 61.3 & 77.8 & 75.2 \\
OC\_SORT~\citep{ocsort} & 63.2 & 78.0 & 77.5 & 62.4 & 75.7 & 76.3 \\
BYTEv2~\citep{bytev2} & 63.6  & 80.6 & 78.9 & 61.4 & 77.3 & 75.6 \\
StrongSORT++~\citep{strongsort} & 64.4 & 79.6 & 79.6 & 62.6 & 73.8 & 77.0 \\
Deep OC\_SORT~\citep{deepocsort} & 64.9 & 79.4 & 80.6 & 63.9 & 75.6 & 79.2 \\
Bot-Sort~\citep{botsort} & 65.0 & 80.5 & 80.2 & 63.3 & 77.8 & 77.5 \\
MotionTrack~\citep{motiontrack_byte_cmc} & 65.1 & 81.1 & 80.1 & 62.8 & 78.0 & 76.5 \\
SparseTrack~\citep{sparsetrack} & 65.1 & 81.0 & 80.1 & 63.5 & 78.1 & 77.6 \\
StrongTBD~\citep{strongtbd} & 65.6 & 81.6 & 80.8 & 63.6 & 78.0 & 77.0 \\
PIA2~\citep{pia2} & 66.0 & 82.2 & 81.1 & 64.7 & 78.5 & 79.0 \\
ImprAsso~\citep{imprasso} & 66.4 & 82.2 & 82.1 & 64.6 & 78.6 & 78.8 \\
SUSHI~\citep{sushi} & 66.5 & 81.1 & 83.1 & 64.3 & 74.3 & 79.8 \\
\hline
DeepMoveSORT-TransFilter (ours) & 63.2 & 78.7 & 77.3 & 60.6 & 73.6 & 74.1 \\
\end{tabular}
\caption{Evaluation results for various state-of-the-art tracking methods on the MOT17 and MOT20 test set. }
\label{tab:results_test_mot17_and_mot20}
\end{table*}

\subsection{Ablation study and experiment discussion}
\label{sec:experiment_discussion}

We perform a vast number of experiments to gain a clear understanding of what exactly enhances tracker performance. We are primarily interested in the benefits for datasets featuring dynamic and non-linear motion. All analyses are conducted on DanceTrack or SportsMOT, with the sole exception of the CMC analysis, which is performed on the MOT17 dataset where it is most beneficial.

\textbf{Tracker component ablation study}. The DeepMoveSORT tracker consists of three main components: a deep learning-based filter, heuristics, and an appearance model. We conduct an ablation study on the DanceTrack and SportsMOT validation sets and analyze the drop in performance of the tracker after removing one or more components. Results can be seen in Table~\ref{tab:dancetrack_ablation} for the DanceTrack dataset and in Table~\ref{tab:sportsmot_ablation} for the SportsMOT dataset. 

For the DanceTrack dataset, we can observe in Table~\ref{tab:dancetrack_ablation} that using the standard KF instead of the TransFilter results in a drastic drop in performance by $4.8\%$ in terms of HOTA, highlighting the importance of a strong motion model for datasets with non-linear motion. Additionally, TransFilter outperforms the improved RNNFilter by $1.0\%$ in HOTA. Removing any heuristic from the tracker logic lowers performance by up to $3.4\%$ in HOTA. The ATCM heuristic proves to be the most beneficial heuristic in this case, mainly due to crowded scenes with many occlusions. Additionally, removing the appearance model results in a performance decrease of $4.5\%$ in HOTA. Based on the results from Table~\ref{tab:dancetrack_ablation}, we conclude that all components are important, with the order of importance being: motion model, appearance model, and heuristics. One should note that HPC and DT-IoU are used for motion-based association, meaning their effectiveness is enhanced by the quality of the motion model.

\begin{table*}%[H]
\centering
\small
\begin{tabular}{ccccc|ccc}
Filter & HVC & DT-IoU & ATCM & Appearance & HOTA & IDF1 & MOTA \\
\hline
TransFilter & Yes & Yes & Yes & Yes & \textbf{62.2} & \textbf{65.3} & 90.9 \\
\underline{KF} & Yes & Yes & Yes & Yes & 57.4 & 59.0 & \textbf{91.1} \\
\underline{RNNFilter} & Yes & Yes & Yes & Yes & 61.2 & 63.7 & 90.9 \\
TransFilter & \underline{No} & Yes & Yes & Yes & 60.5 & 63.9 & 90.8 \\
TransFilter & Yes & \underline{No} & Yes & Yes & 59.7 & 63.7 & 90.8 \\
TransFilter & Yes & Yes & \underline{No} & Yes & 58.6 & 61.2 & 90.8 \\
TransFilter & Yes & Yes & Yes & \underline{No} & 57.5 & 60.9 & 90.7 \\
TransFilter & \underline{No} & \underline{No} & \underline{No} & Yes & 57.7 & 59.9 & 90.6 \\

\end{tabular}
\caption{Ablation study on the DanceTrack validation set for the tracker's filter, heuristics, and appearance-based association. Changes compared to the default settings (i.e.\ first row) are underlined.}
\label{tab:dancetrack_ablation}
\end{table*}

For the SportsMOT dataset, as shown in Table~\ref{tab:sportsmot_ablation}, using the standard KF instead of the TransFilter results in a performance drop of $2.5\%$ in HOTA, while using the improved RNNFilter yields similar performance to the TransFilter. Since the scenes are less crowded and the appearance model is very effective, the benefits of the proposed heuristics are less significant. The DT-IoU is most important, primarily due to the bounding box expansion, which is vital on SportsMOT with very fast movements. If no heuristics are used, the performance drops by $2.0\%$ in HOTA. Finally, removing the appearance association leads to a drastic decrease in performance by $6.6\%$ in HOTA. The component importance order is: appearance model, motion model, and heuristics.

\begin{table*}%[H]
\centering
\small
\begin{tabular}{ccccc|ccc}
Filter & HVC & DT-IoU & ATCM & Appearance & HOTA & IDF1 & MOTA \\
\hline
TransFilter & Yes & Yes & Yes & Yes & \textbf{72.6} & \textbf{78.9} & 91.0 \\
\underline{KF} & Yes & Yes & Yes & Yes & 70.1 & 74.7 & \textbf{91.2} \\
\underline{RNNFilter} & Yes & Yes & Yes & Yes & 72.4 & \textbf{79.0} & 90.8 \\
TransFilter & \underline{No} & Yes & Yes & Yes & \textbf{72.6} & 78.9 & 91.0 \\
TransFilter & Yes & \underline{No} & Yes & Yes & 70.7 & 75.8 & 91.0 \\
TransFilter & Yes & Yes & \underline{No} & Yes & 72.4 & 78.6 & 91.0 \\
TransFilter & Yes & Yes & Yes & \underline{No} & 66.0 & 70.8 & 90.8 \\
TransFilter & \underline{No} & \underline{No} & \underline{No} & Yes & 70.6 & 75.8 & 91.0 \\

\end{tabular}
\caption{Ablation study on the SportsMOT validation set for the tracker's filter, heuristics, and appearance-based association. Changes compared to the default settings (i.e.\ first row) are underlined.}
\label{tab:sportsmot_ablation}
\end{table*}

\textbf{Measurement buffering algorithms comparison}. We compare the MoveSORT's measurement buffering algorithm explained in Section~\ref{sec:background_movesort} and our proposed algorithm for both TransFilter and RNNFilter on the DanceTrack validation set. The results can be seen in Table~\ref{tab:measurements_buffer_algorithm}. The proposed algorithm enhances the performance of the TransFilter, showing improvements of $0.8\%$ in terms of HOTA and $1.4\%$ in terms of IDF1. For the RNNFilter, the algorithm improves performance by $1.2\%$ in HOTA and $1.4\%$ in IDF1. These results demonstrate the generality and efficiency of the proposed measurement buffer method.

\begin{table*}%[H]
\centering
\small
\begin{tabular}{ll|ccc}
Filter & Buffering Algorithm & HOTA & IDF1 & MOTA \\
\hline
TransFilter & MoveSORT & 61.4 & 63.9 & 90.9 \\
TransFilter & DeepMoveSORT & \textbf{62.2} & \textbf{65.3} & 90.9 \\
\hline
RNNFilter & MoveSORT & 60.0 & 62.3 & 90.9 \\
RNNFilter & DeepMoveSORT & \textbf{61.2} & \textbf{63.7} & 90.9 \\
\end{tabular}
\caption{Measurement buffering algorithm comparison between MoveSORT's algorithm and ours on the DanceTrack validation set.}
\label{tab:measurements_buffer_algorithm}
\end{table*}

\textbf{Positional encoding comparison}. We compare our proposed RPE encoding with the standard positional encoding when used with the TransFilter model. The results presented in Table~\ref{tab:results_posenc} indicate that using RPE instead of the standard positional encoding enhances performance by $2.9\%$ in terms of HOTA, $4.6\%$ in terms of IDF1, and $0.1\%$ in terms of MOTA. This suggests that carefully designed positional encoding is very important for transformer-based motion prediction when combined with the measurement buffer during inference in object tracking.

\begin{table*}%[H]
\centering
\small
\begin{tabular}{l|ccc}
Positional encoding & HOTA & IDF1 & MOTA \\
\hline
RPE & \textbf{62.2} & \textbf{65.3} & 90.9 \\
standard & 59.3 & 60.7 & 90.8 \\
\end{tabular}
\caption{TransFilter's positional encoding comparison on the DanceTrack validation set.}
\label{tab:results_posenc}
\end{table*}

\textbf{Training loss function comparison}. We analyze the effects of using different loss functions during the training of the TransFilter model. The results for models trained with MSE and Huber loss are presented in Table~\ref{tab:results_loss_funciton}. We observe an improvement of $0.8\%$ in terms of HOTA, $1.5\%$ in terms of IDF1, and no improvement in terms of MOTA when using Huber instead of MSE loss. This indicates that using a loss function that is robust to outliers is beneficial for training more accurate end-to-end filters.

\begin{table*}%[H]
\centering
\small
\begin{tabular}{l|ccc}
Loss functions & HOTA & IDF1 & MOTA \\
\hline
Huber & \textbf{62.2} & \textbf{65.3} & \textbf{90.9} \\
MSE & 61.3 & 63.8 & 90.9 \\
\end{tabular}
\caption{TransFilter's training loss function comparison on the DanceTrack validation set.}
\label{tab:results_loss_funciton}
\end{table*}

\textbf{HPC heuristic ablation study}. We investigate the usefulness of the object's bounding box height and vertical position cues (HPC heuristic) during the association step in object tracking. Additionally, we compare the proposed heuristic with the existing hybrid association method, i.e.\ combining IoU and L1 distance, which was described in Section~\ref{sec:background_movesort}. For this analysis, SORT tracker is used in combination with the TransFilter model. SORT is convenient for this analysis as it allows the isolation of all other factors, such as appearance features and other heuristics. We present the analysis results in Table~\ref{tab:results_ablation_hpc}, where we observe that the tracker with the proposed HPC heuristic outperforms all other investigated variants. Specifically, the HPC heuristic outperforms the hybrid association method, which additionally uses the object's bounding box width and horizontal position, by $2.6\%$ in terms of HOTA and $3.1\%$ in terms of IDF1. Furthermore, we observe that tracking performance decreases by $1.7\%$ in terms of HOTA when the object's height is ignored during association and by $1.8\%$ in terms of HOTA when the object's vertical position is ignored. This proves that both object's height and vertical position present good cues that should be used during the association step.

\begin{table*}%[H]
\centering
\small
\begin{tabular}{l|ccc}
Association method & HOTA & IDF1 & MOTA \\
\hline
IoU & 51.9 & 90.1 & 51.6 \\
IoU and L1 (hybrid association) & 53.7 & 53.9 & \textbf{90.2} \\
HPC & \textbf{56.2} & \textbf{57.0} & \textbf{90.2} \\
HPC (height only) & 55.3 & 56.1 & 90.0 \\
HPC (y-axis position only) & 55.2 & 56.1 & \textbf{90.2} \\
\end{tabular}
\caption{Comparison between HPC variants and other similar association methods on the DanceTrack validation set. The SORT tracker with the TransFilter motion model was used for all experiments.}
\label{tab:results_ablation_hpc}
\end{table*}

\textbf{Comparison between TCM and ATCM}. We perform a comparison between the track confidence modelling (TCM) approach which uses the constant measurement noise approximation (see Section.~\ref{sec:background_essential_compoents}) and our proposed variant---ATCM, which uses the adaptive measurement noise approximation. We employ the standard ByteTrack tracking method with the TransFilter model to isolate effects from other heuristics and association methods. Note that we use ByteTrack instead of SORT, as the original TCM was proposed for the ByteTrack tracker. Comparison results can be seen in Table~\ref{tab:results_ablation_atcm}, where we observe that ATCM outperforms TCM by $1.0\%$ in terms of HOTA, $0.9\%$ in terms of IDF1, and $0.1\%$ in terms of MOTA. These results demonstrate the effectiveness of our proposed simple modification to the TCM heuristic, where we consider object detection measurements more noisy in case when the confidence output is low.

\begin{table*}%[H]
\centering
\small
\begin{tabular}{l|ccc}
Association method & HOTA & IDF1 & MOTA \\
\hline
IoU and TCM & 53.2 & 55.1 & 90.4 \\
IoU and ATCM & \textbf{54.2} & \textbf{56.0} & \textbf{90.5} \\
\end{tabular}
\caption{Comparison between TCM and ATCM association methods on DanceTack validation set. ByteTrack tracker with the TransFilter motion model was used for all experiments.}
\label{tab:results_ablation_atcm}
\end{table*}

\textbf{CMC efficiency.} We verify the effectiveness of the CMC on the MOT17 dataset, where it is most beneficial. When we applied CMC to other datasets, we observed no significant performance improvements. Since CMC is not a cheap operation in terms of compute, we chose not to use it when it is not necessary. We perform evaluation for the KF, TransFilter, and improved RNNFilter with and without using the CMC. The results can be seen in Table~\ref{tab:results_mot17_cmc_ablation}. In terms of HOTA, using the CMC improves the performance of the KF by $1.0\%$, the TransFilter by $1.7\%$, and the improved RNNFilter by $0.8\%$. This concludes that CMC is fully compatible with the end-to-end learnable filters and further boosts the tracker performance in scenes where the camera motion is not predictable.

\begin{table*}%[H]
\centering
\small
\begin{tabular}{cc|ccc}
Filter & CMC & HOTA & IDF1 & MOTA \\
\hline
KF & No & 69.3 & 80.8 & 79.4 \\
KF & Yes & \textbf{70.3} & \textbf{82.9} & \textbf{80.3} \\
\hline
TransFilter & No & 68.4 & 80.0 & 78.6 \\
TransFilter & Yes & \textbf{70.1} & \textbf{82.8} & \textbf{80.1} \\
\hline
RNNFilter & No & 68.8 & 80.4 & 78.9 \\
RNNFilter & Yes & \textbf{69.6} & \textbf{81.9} & \textbf{79.7} \\
\end{tabular}
\caption{Analysis of CMC benefits for the KF, TransFilter and RNNFilter on the MOT17 validation set.}
\label{tab:results_mot17_cmc_ablation}
\end{table*}

\textbf{Comparing TransFilter to KF across multiple tracking methods}. We compare the KF and TransFilter on the DanceTrack and SportsMOT datasets across different trackers to validate that the benefits of replacing the KF with an end-to-end filter are not tied to a particular algorithm. We consider SORT, SORT-ReID, ByteTrack, ByteTrack-ReID, and DeepMoveSORT. The SORT-ReID and ByteTrack-ReID are trackers that have been extended with a ReID model and appearance association. Results of this analysis can be seen in Table~\ref{tab:results_difference_trackers}, where it is evident that the TransFilter outperforms the KF when used with any tracker. We note that this difference is smaller for trackers with ReID models, which is expected since, to some extent, we can compensate for bad motion prediction by high quality appearance-based association.

\begin{table*}%[H]
\centering
\small
\begin{tabular}{l|ccc|ccc}
& \multicolumn{3}{c|}{DanceTrack} & \multicolumn{3}{c}{SportsMOT} \\
\hline
Tracker & HOTA & IDF1 & MOTA & HOTA & IDF1 & MOTA \\
\hline
SORT-KF & 47.7 & 42.6 & 87.8 & 62.5 & 64.0 & 90.3 \\
SORT-TF & \textbf{51.9} & \textbf{51.6} & \textbf{90.1} & \textbf{62.9} & \textbf{65.4} & \textbf{90.5} \\
\hline
SORT-ReID-KF & 56.6 & 56.9 & 90.5 & 68.5 & 71.5 & 90.9 \\
SORT-ReID-TF & \textbf{57.2} & \textbf{58.7} & \textbf{90.6} & \textbf{68.6} & \textbf{72.3} & \textbf{90.9} \\
\hline
ByteTrack-KF & 52.8 & 52.9 & \textbf{90.6} & 64.2 & 66.9 & \textbf{90.7} \\
ByteTrack-TF & \textbf{53.8} & \textbf{54.5} & \textbf{90.6} & \textbf{65.1} & \textbf{69.0} & \textbf{90.7} \\
\hline
ByteTrack-ReID-KF & 58.2 & 58.4 & 90.9 & 68.8 & 72.8 & \textbf{91.0} \\
ByteTrack-ReID-TF & \textbf{59.0} & \textbf{61.1} & \textbf{91.0} & \textbf{70.6} & \textbf{75.8} & \textbf{91.0} \\
\hline
DeepMoveSORT-KF & 57.4 & 59.0 & \textbf{91.1} & 70.1 & 74.7 & \textbf{91.2} \\
DeepMoveSORT-TF & \textbf{62.2} & \textbf{65.3} & 90.9 & \textbf{72.6} & \textbf{78.9} & 91.0 \\

\end{tabular}
\caption{Effects of replacing KF with TransFilter (TF) for different trackers on the DanceTrack and SportsMOT validation datasets.}
\label{tab:results_difference_trackers}
\end{table*}

\begin{figure*}%[H]
  \centering
  \includegraphics[width=1.0\linewidth]{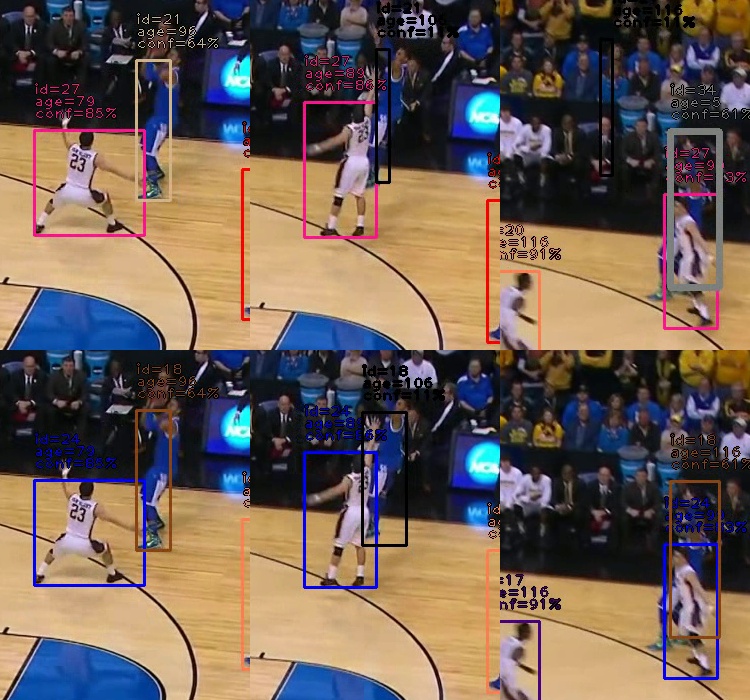}
  \caption{Visual comparison between DeepMoveSORT-KF (first row) and DeepMoveSORT-TransFilter (second row) on SportsMOT \textit{v\_00HRwkvvjtQ\_c007}. We present snapshots of a basketball player shooting a ball at frames $95$, $105$, and $115$, displayed in each column respectively. The detector fails to detect the player during the jump, indicating that the player is occluded (predicted bounding boxes of occluded tracks are shown in black). It is observed that the DeepMoveSORT-KF loses the player due to poor and degenerate bounding box predictions, particularly in terms of aspect ratio. After the occlusion, a new track for the same player is initialized (bounding box with thick gray color) as the previous one was lost. In contrast, the DeepMoveSORT-TransFilter maintains accurate tracking, with high-quality predictions and realistic bounding box aspect ratios.}
  \label{fig:kf_vs_tf_basketball_example}
\end{figure*}

\begin{figure*}%[H]
  \centering
  \includegraphics[width=1.0\linewidth]{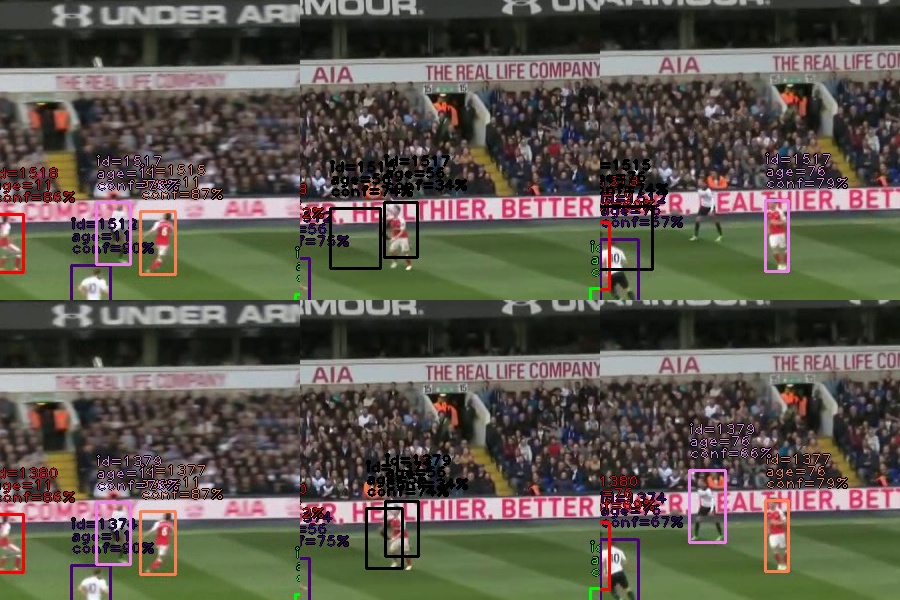}
  \caption{Visual comparison between DeepMoveSORT-KF (first row) and DeepMoveSORT-TransFilter (second row) on SportsMOT \textit{v\_ITo3sCnpw\_k\_c011}. We present snapshots of two football players rapidly changing their direction at frames $10$, $55$, and $75$, displayed in each column respectively. The detector fails to detect both players around frame $55$, indicating that the players are occluded (predicted bounding boxes of occluded tracks are shown in black). It is observed that the DeepMoveSORT-KF performs an identity switch between those two players and additionally loses one player. In contrast, the DeepMoveSORT-TransFilter maintains both players' identity.}
  \label{fig:kf_vs_tf_football_example}
\end{figure*}

We also perform a visual analysis for both DeepMoveSORT-KF and DeepMoveSORT-TF trackers. We observe that both the KF and TransFilter perform similarly in easy cases, such as when an unaccompanied athlete is walking or running without changing direction. However, TransFilter outperforms KF in more challenging scenarios, such as when an athlete is jumping or changing direction in a crowd. Additionally, TransFilter does not predict degenerate bounding boxes with unrealistic aspect ratios, unlike KF. This is expected, as KF is not able to learn this from the data. We present two such examples in Figure~\ref{fig:kf_vs_tf_basketball_example} and Figure~\ref{fig:kf_vs_tf_football_example}.

\textbf{Noise Filtering Ablation}. We investigate the efficiency of noise filtering by evaluating the SORT tracker with the TransFilter model in two scenarios: using only the motion model (filtering disabled) and using both the motion model and noise filtering (filtering enabled). For these experiments, we simulate localization error noise by adding Gaussian noise to all oracle detection coordinates, with the standard deviation proportional to the bounding box scale. Since the filtering is performed to improve bounding box localization in object detection, we use the DetA metric as it predominantly relates to the quality of detections. All experiments are repeated five times, and the metrics are averaged.

From Table~\ref{tab:noise_filtering} we observe that when using oracle detections (perfect detector), filtering offers no benefits since the observations are already noiseless. When noise is added to bounding box coordinates, tracker performance drops significantly in terms of DetA. However, with filtering enabled, we can observe that this drop is partially mitigated. Stronger noise mitigation is not expected, as noise filtering cannot compensate for a fundamentally flawed detector, but it can make a noticeable difference.

In cases where the YOLOX object detection model is used, we note that the benefits of noise filtering are not pronounced, showing no relevant difference, i.e.\ less than $0.1\%$, in DetA. This is due to the fact that the YOLOX model already performs localization effectively on the DanceTrack dataset. Larger benefits are expected with less accurate detectors.

\begin{table*}%[H]
\centering
\small
\begin{tabular}{c|cc}
noise magnitude & filtering disabled & filtering enabled \\
\hline
0.000 & 98.4 & 98.3 \\
0.025 & 89.6 & 90.3 \\
0.050 & 80.7 & 82.6 \\
0.100 & 65.1 & 70.3 \\

\end{tabular}
\caption{Evaluation of DetA results for different magnitudes of the augmentation noise on the DanceTrack validation set. All experiments are repeated 5 times and the metrics are averaged.}
\label{tab:noise_filtering}
\end{table*}

\textbf{Computational Efficiency}. We measure the throughput of each method in frames per second (FPS), with higher FPS indicating faster performance. We performed inference for the TransFilter, RNNFilter, and KF with a batch size of 256, i.e.\ processing 256 objects at a time, using a single core of an Intel i7-12700K processor and a single NVIDIA GeForce RTX 3070 GPU. We considered inference time for 1-step (no occlusion) and 30-step (long occlusion) predictions. The inference speed measurement included pre-processing, model prediction, and post-processing stages. The TransFilter model relies on the input trajectory length of 10, while the RNNFilter model uses a length of 30. 

The speed comparison of all filters can be seen in Table~\ref{tab:computational_efficiency}. Based on these results, we observe that for the 1-step prediction, the KF is up to 30 times faster than the TransFilter, while the TransFilter performs inference slightly faster than the KF during long occlusions, i.e.\ 30-step prediction. The RNNFilter with its default setting is the slowest. It becomes comparable to the TransFilter when the input trajectory history length is reduced, but at the expense of its tracking performance. We note that these results were obtained using the PyTorch~\citep{pytorch} framework without any optimizations for inference time. In practice, all frame rates could be higher if model weights were converted to a more efficient format.

From the results in Table~\ref{tab:computational_efficiency}, we observe that inference during occlusions is noticeably faster for the TransFilter and RNNFilter than for the KF. This is because in these models the input trajectory does not need to pass through the encoder once for each generated prediction. Instead, all predictions are made based on one pass of the input trajectory through the encoder. Additionally, the TransFilter model performs inference only at the beginning of the object's occlusion, as its architecture is designed for single-shot, multi-step predictions. However, this speedup during occlusions does not apply when using the old measurement buffering algorithm proposed by MoveSORT, as it updates the measurement buffer at every step during the occlusion. When using MoveSORT's measurement buffering algorithm, the 30-step prediction inference speed is similar to the 1-step prediction speed, which is significantly slower. Therefore, our design of the proposed measurement buffering algorithm yields speed benefits, too.

\begin{table*}%[H]
\centering
\small
\begin{tabular}{l|cc|cc}
 & \multicolumn{2}{c|}{full prediction} & \multicolumn{2}{c}{model prediction} \\
Filter & 1-step & 30-step & 1-step & 30-step \\
\hline
TransFilter (10) & 160 & \textbf{4770} & 174 & \textbf{12570} \\
RNNFilter (30) & 87 & 1983 & 92 & 2121\\
RNNFilter (10) & 165 & 3378 & 180 & 3711 \\
KF & \textbf{4522} & 451 & \textbf{4522} & 4512 

\end{tabular}
\caption{Computational efficiency comparison in terms of FPS (Frames Per Second) for RNNFilter, TransFilter, and KF filters implemented in PyTorch. Full prediction time includes pre-processing, model prediction, and post-processing. Two scenarios are analyzed: non-occluded object (1-step prediction) and long-term occlusion of the object (30-step prediction). We denote length of the input trajectory in the filter name parentheses.}
\label{tab:computational_efficiency}
\end{table*}

\textbf{Training Filters on Real Measurements Instead of Simulated Ones}. Following~\citep{movesort}, we trained our end-to-end filters on ground truth annotations augmented with noise. This includes adding Gaussian noise to bounding box coordinates to simulate object detection localization errors, and masking bounding boxes to simulate object detection false negative errors. Filters trained in this way are not specialized for any specific object detector. However, in this work, we also considered an alternative approach, which involves training models on real object detection measurements that are used in inference. We performed empirical analysis and based on the results of the experiments described in~\ref{appendix:additional_experimental_analysis}, we concluded that there is no benefit to training on real object detection measurements when using the previously mentioned augmentations during training. This is a beneficial property because we can keep our end-to-end filters independent of a specific object detection model and switch them at will.

\textbf{Using Image Features for Motion Prediction}. We considered extending the TransFilter motion model by combining trajectory features with image features. The motivation for these experiments was the hypothesis that an object's motion depends on the context, which may not be fully contained in the trajectory history but could be discerned from the surroundings in the image. An example of such context would be the location of a sport ball, which suggests the players' future movement. We performed empirical analysis and based on the results of the experiments described in~\ref{appendix:additional_experimental_analysis}, we concluded that combining trajectory and image features with the proposed architectures does not provide any benefits in terms of motion prediction accuracy. We speculate that most of the relevant information for short-term trajectory prediction is already included in trajectory features.

\section{Related work}
\label{sec:related_work}

In this section, we compare various aspects of our approach with the related work. Firstly, we compare our proposed end-to-end filter, TransFilter, with existing filters and motion models used for object tracking. Secondly, we discuss the modifications we applied to the measurement buffering algorithm that are related to all end-to-end filters. Lastly, we examine the related work corresponding to each of our proposed heuristic-based association methods: HPC, DT-IoU, and ATCM. 

\textbf{TransFilter}. The TransFilter architecture can be seen as an adaptation of the standard transformer used for natural language processing~\citep{attention_is_all_you_need} to the end-to-end filtering framework. In this architecture, the transformer encoder is used for motion prediction, while the decoder handles noise filtering. The only modification we applied to the encoder is replacing the standard positional encoding with Reverse Positional Encoding (RPE)~\citep{rpe}. In the case of the decoder, we removed the multi-head attention layer, which is redundant in our scenario, and also excluded the positional encoding. For token embedding we use a MLP applied to trajectory features instead of a linear layer used in the standard transformer architecture.

Compared to the RNNFilter and NODEFilter recursive end-to-end filters, TransFilter is fully transformer-based and does not use GRUs for noise filtering. Additionally, TransFilter is non-recursive in both motion prediction and noise filtering, hence it employs parallel computing during training and inference. Compared to the recursive end-to-end filters, TransFilter does not accumulate errors in multi-step motion prediction.

Compared to the motion model proposed by MotionTrack~\citep{motiontrack}, the TransFilter architecture does not use the Dynamic MLP layer to model channel-level features. Instead, it aligns more closely with the standard transformer architecture. Unlike the MotionTrack, which solely performs motion prediction, TransFilter can also perform noise filtering in scenarios of frequent object localization errors.

\textbf{Measurement Buffering Algorithm}. Compared to the original measurement buffering algorithm mentioned in Section~\ref{sec:background_movesort}, we avoid updating the buffer during occlusions. This approach allows us to retain sufficient measurements in the buffer without using any history extension logic proposed in the original algorithm. However, in contrast, we rapidly update the buffer after the occlusion ends to keep the measurements up to date. Additionally, we have defined a method for updating the buffer with CMC affine transformations in scenarios where camera motion is only predictable through image features. We note that all our proposed modifications apply not only to the end-to-end filters but also to the deep learning-based Bayesian filters proposed by~\citep{movesort}.

\textbf{HPC association}. Our proposed HPC (Horizontal Perspective Cues) association method is a modified and tuned version of the MoveSORT's hybrid association method explained in Section~\ref{sec:background_movesort}. The primary difference is that we ignore the object’s bounding box width and horizontal position, focusing instead only on height and vertical position. Additionally, we apply separate weights to the distances for height and vertical position, which adds flexibility to this heuristic.

HPC can also be compared to other association methods that use heuristics related to the object's bounding box size and position. SparseTrack~\citep{sparsetrack} uses cascaded associations by sorting all bounding boxes based on their position on the vertical axis, which they refer to as ``approximated depth''. In contrast to SparseTrack, we avoid using cascaded association and perform association for all objects at once. Moreover, we consider the object's height during the association step. Hybrid-SORT~\citep{hybridsort} employs height modulated IoU (HMIoU) which is equal to the bounding box IoU multiplied by the vertical axis IoU. This approach effectively incorporates height differences into the association. Compared to HMIoU, our HPC heuristic additionally considers the object’s vertical position and can be used independently of the IoU association. 

\textbf{DT-IoU association}. Our proposed DT-IoU (Decay Threshold IoU) association method can be seen as a generalization of the standard IoU association method used in modern tracking-by-detection algorithms~\citep{bytetrack, botsort, sparsetrack, motiontrack}. When the upper and lower IoU bounds of the DT-IoU are equal and expansion is not employed, this method is equivalent to the standard IoU association method. DT-IoU method is more flexible as the IoU threshold reduces when the object is occluded, i.e.\ when we are less certain about the object's position.

The DT-IoU supports bounding box expansion in cases where the object's movements are very rapid compared to its size, i.e.\ in cases where the overlap between bounding boxes from consecutive frames is small or non existent. This expansion was inspired by previous work on the CBIoU~\citep{cbiou} and Deep-EIoU~\citep{deep_eiou} trackers. CBIoU and Deep-IoU, which generalizes CBIoU, perform cascaded association by expanding the sizes of predicted and detection bounding boxes in subsequent cascades when initial associations are unsuccessful. Compared to these methods, we use a fixed expansion rate without cascaded association. By avoiding the cascade association, our approach might perform worse in isolation, but it remains flexible in scenarios where it could be combined with other heuristics.

\textbf{ATCM Association}. Our proposed ATCM (Adaptive Track Confidence Modeling) association method is a modified version of the TCM heuristic proposed by Hybrid-SORT~\citep{hybridsort}. In the original TCM method, which employs the KF for confidence modeling, a constant value is used for approximating the measurement (confidence) likelihood. Instead, we propose an adaptive confidence approximation in which the confidence uncertainty is inversely proportional to the object detection confidence. In other words, the more confident the object detection model is, the more the KF trusts those measurements.

\section{Conclusion}
\label{sec:conclusion}

In this paper, we introduced a novel tracking method---DeepMoveSORT, designed to address the challenges in multi-object tracking, particularly in scenarios involving dynamic, non-linear motion. DeepMoveSORT combines a new learnable filter with appearance similarity and a rich set of new heuristics to achieve state-of-the-art tracking performance. Specifically, we proposed TransFilter, a transformer-based end-to-end learnable filter that performs both motion prediction and noise filtering. Moreover, we enhanced the filters' performance in two significant ways. First, we refined the measurement buffering algorithm, which maintains the object's motion history used for the model's prediction. This refinement leads to increased speed and accuracy during inference. We also incorporated camera motion compensation to update the measurement buffer in case a non-static camera is used. To reduce the number of identity switches, we proposed three heuristics: HPC, which uses cues from the object's position and shape; ATCM, which uses detection confidence as a metric for association; and DT-IoU, which generalizes the standard IoU and accounts for the loss of certainty about an object's position during occlusion. We performed a thorough evaluation to establish that DeepMoveSORT tracker surpasses the best performing trackers on datasets featuring dynamic, non-linear motion, such as DanceTrack, SportsMOT, and SoccerNet. Moreover, we conducted a detailed ablation study to evaluate the contribution of each proposed improvement. Our main observations, in order of importance, can be summed up as follows:
\begin{itemize}
\item Using a learnable filter instead of the KF, and using appearance similarity for association has proven highly beneficial on all datasets. 
\item Choosing the right learnable filter architecture is important. Using TransFilter provides a strong boost on the DanceTrack dataset, but RNNFilter remains the most effective on the SportsMOT dataset. However, we noticed that this difference in accuracy between the learnable filters is much smaller compared to the difference between a learnable filter and the KF. Therefore, TransFilter might be considered as the default option.
\item Using our proposed heuristics DT-IoU, HPC, and ATCM is crucial for strong performance on crowded datasets. We observed that DT-IoU contributes the most across all datasets, while HPC and ATCM have significant importance on DanceTrack, which is the most crowded dataset.
\end{itemize}
To conclude, our work provides both a high performing tracker for non-linear motion and the insights on good engineering practices when building one.

\section*{Acknowledgements}

Predrag Tadić acknowledges the financial support of the Ministry of Science, Technological Development and Innovation of the Republic of Serbia under contract number 451-03-65/2024-03/200103.

\section*{Declaration of competing interests}

The authors declare that they have no known competing financial interests or personal relationships that could have appeared to influence the work reported in this paper.

\section*{Data availability}

All datasets used in this study are publicly available.

\appendix

\section{Filter's trajectory features}
\label{appendix:trajectory_features}

For trajectory input features, we adhere to the methodology outlined in~\citep{movesort}, combining three types of trajectory features: absolute bounding box coordinates, which include the precise position of the bounding box within the image and the object's size; {\em standardized scaled first order differences}---the coordinate differences between consecutive frames providing a rough approximation of velocities for each coordinate; {\em standardized scaled relative to last observation coordinates}---bounding box coordinates relative to the last observed bounding box, serving as a translation-invariant version of the absolute coordinates, and time relative to the last observation. The sole deviation from~\citep{movesort} is the scaling of coordinates relative to the last observation, now adjusted based on the time difference between bounding box being transformed and the last observed bounding box. For trajectory target features, we also employ the approach of~\citep{movesort}, using {\em standardized relative to last observation coordinates} alongside time relative to the last observation.

\section{Detailed experimental setup}
\label{appendix:detailed_experimental_setup}

This section provides comprehensive details about the object detection, ReID, and filter models, which are described in the subsequent sections.

\subsection{Object detection}
\label{appendix:object_detection}

This section includes information about each object detection model grouped by datasets.

\textbf{DanceTrack}. For the DanceTrack evaluation, we use the publicly available DanceTrack YOLOX object detector~\citep{yolox}. This detector is trained only on the training set.

\textbf{SportsMOT and SoccerNet}. For the SportsMOT test set evaluation, we use YOLOX detector fine-tuned for the Deep-EIoU tracker~\citep{deep_eiou}. This detector was trained using both the training and validation sets. For ablation study and hyper-parameter tuning, we fine-tuned YOLOv8~\citep{yolov8} pretrained on the COCO dataset~\citep{coco}. The same YOLOX model is also used for the SoccerNet dataset benchmark.

\textbf{MOTChallenge}. We use ByteTrack's YOLOX for all MOTChallenge evaluation and benchmarks. We follow the procedure defined by the MoveSORT~\cite[Appendix B.2,~Appendix B.3]{movesort}.

\subsection{Appearance model}
\label{appendix:reid}

This section includes information about each ReID model. For all evaluations we use the FastReid's BoT(SBS50)~\citep{fastreid, bot} architecture.

\textbf{DanceTrack}. For the DanceTrack evaluation, we use the model fine-tuned by the Hybrid-SORT~\citep{hybridsort}. This model was trained on both DanceTrack and CUHKSYSU~\citep{cuhksysu} datasets.

\textbf{SportsMOT, SoccerNet}. For the SportsMOT, we fine-tune a model starting from checkpoint trained on the Market1501~\citep{market1501} dataset. We use the same model for SoccerNet without any fine-tuning.

\textbf{MOTChallenge}. We use Bot-SORT's models for all MOTChallenge evaluation and benchmarks. Separate models are fine-tuned by Bot-SORT~\citep{botsort} for MOT17 and MOT20.

\subsection{TransFilter and RNNFilter technical details}
\label{appendix:training_filters}

In this section, we detail the technical aspects related to the architectures and training procedures of the TransFilter and improved RNNFilter.

\textbf{Architecture}. For all datasets, the history length is set to 10 for TransFilter and 30 for RNNFilter. The only exception is DanceTrack, in which case we use history length 10 for RNNFilter. On DanceTrack, MOT17, and MOT20, both models predict up to 30 steps ahead, while for SportsMOT (and SoccerNet), the prediction extends to 60 steps. All models are configured with a feature channel width of 256. We use 6 stacked encoder layers for the TransFilter model and a single decoder layer. Unless specified otherwise, all linear layers are followed by layer normalization~\citep{layer_norm}, and then by SiLU~\citep{silu}. 

\textbf{Training}. We employ the AdamW optimizer~\citep{adamw} with an initial learning rate of $5e^{-5}$ for TransFilter and $1e^{-3}$ for RNNFilter. Both models use a weight decay of $1e^{-4}$. The AdamW optimizer settings include a running average coefficient of $0.9$ for gradients and $0.999$ for the squares of gradients. A step learning rate scheduler reduces the learning rate by a factor of $0.1$ every fourth epoch. Additionally, for TransFilter, a linear learning rate warm-up is applied during the first four epochs as recommended by~\citep{attention_is_all_you_need}. For Huber loss, we use delta coefficient equal to $0.5$ on all datasets. For the input trajectory augmentations, we use the same procedure as outlined in the~\cite[A.3]{movesort}. These augmentations include adding Gaussian noise to bounding box coordinates to simulate detector localization errors and removing random trajectory points to simulate detector false negative errors.

\subsection{Tracker hyper-parameters}
\label{appendix:tracker_hyperparameters}

We present the hyper-parameter values for each dataset in Table~\ref{tab:tracker_hyperparameters}. We define three states of object tracking for clearer understanding of the hyper-parameters: an \textit{active track} is the track of an object that was successfully associated with a detection in the last frame; a \textit{lost track} refers to the track of an occluded object; and a \textit{new track} is a potential new object that needs to be confirmed by being successfully associated with detections over few consecutive frames. Hyper-parameters specific to ByteTrack's high detection confidence (first) association are tagged as HA (\textit{High Association}), those related to ByteTrack's low detection confidence (second) association as LA (\textit{Low Association}), and those for new track associations as NA (\textit{New Association}). Explanations for each hyper-parameter can be found in the following list:

\begin{itemize}
    \item \textbf{Detection confidence threshold}: The threshold based on which detected bounding boxes are split for ByteTrack's high detection confidence (first) cascade and low detection confidence (second) cascade. Bounding boxes with confidence lower than $0.1$ are completely ignored.
    \item \textbf{Track max time lost}: Occluded objects are not considered lost, i.e., their tracks are not deleted and can be re-identified until the defined maximum occlusion time is reached.
    \item \textbf{Track initialization time}: New (potential) tracks must be successfully associated for a certain number of consecutive frames to become active (initialization phase), or else they are deleted. This initialization phase is necessary to make tracker robust to object detection false positives.
    \item \textbf{Track initialization confidence}: During the new track initialization phase, detected bounding boxes need to have high confidence; otherwise, the new track is not initialized.
    \item \textbf{Duplicate track IoU threshold}: Remove active or lost tracks if they almost fully overlap with another track. The longer-existing track is kept while the other one is deleted.
    \item \textbf{Apply noise filtering}: Perform noise filtering after the track is associated with a detection.
    \item \textbf{Use CMC}.
    \item \textbf{DT-IoU threshold upper bound}: $IoU_{\text{upper\_bound}}$ defined in Section~\ref{sec:association_methods}.
    \item \textbf{DT-IoU threshold lower bound}: $IoU_{\text{lower\_bound}}$ defined in Section~\ref{sec:association_methods}.
    \item \textbf{DT-IoU threshold decay}: $IoU_{\text{decay}}$ defined in Section~\ref{sec:association_methods}.
    \item \textbf{DT-IoU expansion rate}: Bounding box height and width expansion rate defined in Section~\ref{sec:association_methods}.
    \item \textbf{DT-IoU (IoU) fuse detection score}: If used, the IoU score is multiplied by the detection confidence (inherited from ByteTrack's implementation).
    \item \textbf{DT-IoU weight}: $\lambda_{DT-IoU}$ defined in Section~\ref{sec:association_methods}.
    \item \textbf{ReID weight}: $\lambda_{Appr}$ defined in Section~\ref{sec:association_methods}.
    \item \textbf{ATCM weight}: $\lambda_{ATCM}$ defined in Section~\ref{sec:association_methods}.
    \item \textbf{HPC weight}: $\lambda_{HPC}$ defined in Section~\ref{sec:association_methods}.
    \item \textbf{HPC height weight}: $\lambda_{h}$ defined in Section~\ref{sec:association_methods}.
    \item \textbf{HPC vertical position weight}: $\lambda_{y}$ defined in Section~\ref{sec:association_methods}.
\end{itemize}

\begin{table*}
\tiny
\centering
\begin{tabular}{l|cccccc}
Hyper-parameter & DanceTrack & SportsMOT & SoccerNet & MOT17 & MOT20 \\
\hline
detection confidence threshold & 0.6 & 0.6 & 0.6 & 0.6 & 0.6 \\
track max time lost & 30 & 60 & 60 & 30 & 30 \\
track initialization time & 3 & 3 & 3 & 3 & 3 \\
track initialization confidence & 0.70 & 0.70 & 0.70 & 0.70 & 0.70 \\
duplicate track IoU threshold & 1.00 & 1.00 & 1.00 & 0.70 & 1.00 \\
apply noise filtering & no & no & no & yes & no \\
use CMC & no & no & no & yes & no \\
(HA) DT-IoU threshold upper bound & 0.50 & 0.20 & 0.20 & 0.40 & 0.40 \\
(HA) DT-IoU threshold lower bound & 0.25 & 0.05 & 0.05 & 0.20 & 0.30 \\
(HA) DT-IoU threshold decay & 0.20 & 0.05 & 0.05 & 0.10 & 0.10 \\
(HA) DT-IoU expansion rate & 0.00 & 0.70 & 0.70 & 0.00 & 0.00 \\
(HA) DT-IoU fuse detection score & no & no & no & yes & no \\
(LA) IoU threshold & 0.50 & 0.50 & 0.50 & 0.50 & 0.50 \\
(LA) IoU expansion rate  & 0.00 & 0.35 & 0.35 & 0.00 & 0.00 \\
(LA) IoU fuse detection score & no & no & no & yes & no \\
(NA) IoU threshold & 0.25 & 0.15 & 0.15 & 0.30 & 0.30 \\
(NA) IoU threshold expansion rate & 0.00 & 0.70 & 0.70 & 0.00 & 0.00 \\
(NA) IoU fuse detection score & no & no & no & yes & no \\
(HA) DT-IoU weight & 1.00 & 1.00 & 1.00 & 0.20 & 0.20 \\
(HA) ReID weight & 2.00 & 2.00 & 2.00 & 0.80 & 0.80 \\
(HA) ATCM weight & 1.50 & 0.50 & 0.50 & 0.00 & 0.20 \\
(HA) HPC weight & 2.00 & 1.00 & 1.00 & 1.00 & 0.20 \\
(HA) HPC height weight & 1.00 & 1.00 & 1.00 & 1.00 & 1.00 \\
(HA) HPC vertical position weight & 1.00 & 0.50 & 0.50 & 0.00 & 1.00 \\
(LA) IoU weight & 1.00 & 1.00 & 1.00 & 0.20 & 0.20 \\
(LA) ATCM weight & 1.00 & 1.00 & 1.00 & 0.00 & 0.20 \\
(LA) HPC weight & 2.00 & 1.00 & 1.00 & 0.20 & 0.10 \\
(LA) HPC height weight & 1.00 & 1.00 & 1.00 & 1.00 & 1.00 \\
(LA) HPC vertical position weight & 1.00 & 0.50 & 0.50 & 0.00 & 1.00 \\
(NA) IoU weight & 1.00 & 1.00 & 1.00 & 0.20 & 0.2 \\
(NA) HPC weight & 2.00 & 1.00 & 1.00 & 0.20 & 0.2 \\
(NA) HPC height weight & 1.00 & 1.00 & 1.00 & 1.00 & 1.00 \\
(NA) HPC vertical position weight & 1.00 & 1.00 & 1.00 & 0.00 & 1.00 \\
\end{tabular}
\caption{Tracker hyper-parameter configuration DanceTrack, SportsMOT, SoccerNet, MOT17 and MOT20 datasets.}
\label{tab:tracker_hyperparameters}
\end{table*}

\section{Additional results}

This section includes additional benchmark and experimental analysis results.

\subsection{Additional benchmarks}
\label{appendix:additional_benchmarks}

This section includes additional, less relevant benchmarks.

\textbf{SoccerNet with oracle detections}. We compare performance when oracle detections are used instead of a real object detector, as more results are published with this experimental setup. However, we consider these results to be less relevant. Additionally, we note that we did not further tune any tracker hyper-parameters specifically for the oracle detections. The results can be seen in Table~\ref{tab:results_test_soccernet_oracle}. In this case, DeepMoveSORT still outperforms most methods, including Deep-EIoU, but lags behind CBIoU~\citep{cbiou}.

\begin{table*}%[H]
\centering
\small
\begin{tabular}{l|ccc}
Method & HOTA & DetA & AssA \\
\hline
DeepSORT~\citep{deepocsort} & 69.6 & 82.6 & 58.7 \\
ByteTrack~\citep{bytetrack} & 71.5 & 84.3 & 60.7 \\
Bot-Sort~\citep{botsort} & 77.0 & 93.5 & 63.4 \\
OC\_SORT~\citep{ocsort} & 78.1 & 94.3 & 64.7 \\
Deep-EIoU~\citep{deep_eiou} & 85.4 & \textit{99.2} & 73.6 \\
CBIoU~\citep{cbiou} & \textbf{89.2} & \textbf{99.4} & \textbf{80.0} \\
\hline
DeepMoveSORT-TransFilter (ours) & 86.2 & 98.1 & 75.8 \\
DeepMoveSORT-RNNFilter (ours) & \textit{86.8} & 98.3 & \textit{76.6}
\end{tabular}
\caption{Evaluation results on SoccerNet test with oracle detections. The best results are in bold for each metric, while the second-place results are italicized. The results for the DeepSORT, ByteTrack are taken from~\citep{soccernet}, while Bot-Sort and OC\_SORT are taken from~\citep{deep_eiou}.}
\label{tab:results_test_soccernet_oracle}
\end{table*}

\subsection{Additional experimental analysis}
\label{appendix:additional_experimental_analysis}

This section includes extended set of experiments and analysis that we performed.

\textbf{Training with Mined Detection Bounding Boxes}. We explore the use of real measurements for motion prediction and noise filtering training. We define the process of obtaining these measurements as \textit{detection mining}, and the results are termed \textit{mined detection bounding boxes}. This process involves performing object detection inference on each frame to obtain bounding boxes and assigning track IDs to all high-confidence bounding boxes. We only consided bounding boxes that have confidence at least $0.3$. We employ the Hungarian algorithm to assign track IDs to detection bounding boxes based on their overlap with the ground truth. We define a minimum IoU threshold for matching equal to $0.7$. Any detection bounding box that cannot be matched with a ground truth is discarded. Matched detection bounding boxes are labeled with the corresponding ground truth track ID. Essentially, we generate a dataset similar to an annotated one, albeit with inherent object detection noise. To account for unmatched ground truths, we still use a small proportion of them---approximately $10\%$---to compoensate for the missing detection bounding boxes.

To evaluate the efficacy of the mined dataset, we compare models trained exclusively on ground truths with those trained on mined detections. These comparisons are presented in Table~\ref{tab:results_simulated_vs_detections}. The results indicate that training on mined detections offers no significant advantage over training solely on ground truths, whether or not augmentations such as Gaussian noise and false negatives are employed. We conclude that the augmentations proposed by MoveSORT~\citep{movesort} play a more crucial role.

\begin{table*}%[H]
\centering
\small
\begin{tabular}{cc|ccc}
Inputs & Augmentations & HOTA & IDF1 & MOTA \\
\hline
GT & Yes & \textbf{62.2} & \textbf{65.3} & \textit{90.9} \\
GT & No & 60.6 & 62.7 & 91.0 \\
Detections & Yes & \textit{61.9} & \textit{64.6} & \textbf{91.0} \\
Detections & No & 61.0 & 63.0 & 90.9
\end{tabular}
\caption{Comparison of the DeepMoveSORT-TransFilter performance when using ground truths, i.e.\ oracle detections, or mined object detections combined with trajectory augmentations during training on DanceTrack validation set. The best results are in bold for each metric, while the second-best results are italicized.}
\label{tab:results_simulated_vs_detections}
\end{table*}

\textbf{Combining trajectory and image features}. We explored multiple architectures as extensions of the the TransFilter. Specifically we use multiple encoders for different type of signals (i.e. trajectory, image) and concatenate their vector representation outputs of each encoder before using the prediction head. We considered following architectures:
\begin{itemize}
    \item Combining the trajectory encoder and image encoder with full image view. The last observed frame is used as the image input to the image encoder.
    \item Combining the trajectory encoder and image encoder with expanded bounding box crop instead of the full image view. The crop from the last observed frame is used as the image input to the image encoder.
    \item Combining the trajectory encoder, image encoder, and image motion encoder. The last two expanded bounding box crops from the last observed frame are used. The image encoder processes the last crop, while the image motion encoder processes the difference between the last two crops, i.e., absolute pixel difference.
\end{itemize}

We evaluated all proposed architectures on the SportsMOT validation set. To better understand the usefulness of various image features, we included architectures that do not use any trajectory features. According to the results shown in Table~\ref{tab:image_features}, we conclude that relying solely on image features is insufficient for accurate motion prediction, as these models drastically underperform. Additionally, while combining trajectory and image features yields improved results, they still do not outperform the trajectory-only architecture.

\begin{table*}%[H]
\centering
\small
\begin{tabular}{l|ccc}
Input features & HOTA & IDF1 & MOTA \\
\hline
trajectory only & \textbf{72.5} & \textbf{78.7} & \textit{91.0} \\
image only (full view) & 67.1 & 71.5 & 90.2 \\
image only (expanded box crop) & 69.3 & 74.8 & 90.4 \\
image and image motion (expanded box crop) & 69.6 & 75.2 & 90.4 \\
trajectory and image (expanded box crop) & \textit{71.9} & \textit{78.0} & \textbf{91.1} \\
trajectory, image and image motion (expanded box crop) & 71.6 & 77.4 & 91.0 \\
\end{tabular}
\caption{Comparison between DeepMoveSORT with various motion model input features on SportsMOT validation set. The best results are in bold for each metric, while the second-best results are italicized.}
\label{tab:image_features}
\end{table*}

\clearpage

\bibliographystyle{elsarticle-harv}
\bibliography{main}

\end{document}